\documentclass[10pt,journal,compsoc]{IEEEtran}
\usepackage{underscore}
\usepackage{adjustbox}
\usepackage{hyperref}
\usepackage{amssymb}
\usepackage{graphicx}
\usepackage{multirow}
\usepackage{bbding}
\usepackage{cancel}
\usepackage{url}
\usepackage{graphicx}
\usepackage{color}
\usepackage[ruled]{algorithm2e}
\usepackage{algorithm2e,setspace}
\usepackage[noend]{algpseudocode}
\usepackage{stfloats}
\usepackage{multirow}
\usepackage{makecell}

\usepackage{bm}
\usepackage[table,xcdraw]{xcolor}
\usepackage[normalem]{ulem}
\useunder{\uline}{\ul}{}
\ifCLASSOPTIONcompsoc
  \usepackage[nocompress]{cite}
\else
  \usepackage{cite}
\fi
\usepackage{ragged2e}

\usepackage{float}  
\usepackage{lipsum}


\ifCLASSINFOpdf
\else
\fi
\begin{document}
%\title{Knowledge Distillation Meets Federated Multi-task Learning in Mobile Edge Computing}

\title{FedICT: Federated Multi-task Distillation for Multi-access Edge Computing}

\author{Zhiyuan~Wu,~\IEEEmembership{Member,~IEEE,}
        Sheng~Sun,
        Yuwei~Wang,~\IEEEmembership{Member,~IEEE,}\\
        Min~Liu,~\IEEEmembership{Senior~Member,~IEEE,}
        Quyang Pan,
        Xuefeng~Jiang,~and~Bo~Gao,~\IEEEmembership{Member,~IEEE}
\IEEEcompsocitemizethanks{\IEEEcompsocthanksitem Zhiyuan Wu and Xuefeng Jiang are with the Institute of Computing Technology, Chinese Academy of Sciences, Beijing, China, and also with the University of Chinese Academy of Sciences, Beijing, China.\protect\\
E-mail: \{wuzhiyuan22s, jiangxuefeng21b\}@ict.ac.cn.
\IEEEcompsocthanksitem Sheng Sun, Yuwei Wang and Quyang Pan are with the Institute of Computing Technology, Chinese Academy of Sciences, Beijing, China.
\protect\\
E-mail: \{sunsheng, ywwang\}@ict.ac.cn, lightinshadow1110@gmail.com.
\IEEEcompsocthanksitem Min Liu is with the Institute of Computing Technology, Chinese Academy of Sciences, Beijing, China, and also with the Zhongguancun Laboratory, Beijing, China.
\protect\\
E-mail: liumin@ict.ac.cn.
\IEEEcompsocthanksitem Bo Gao is with the School of Computer and Information Technology, and the Engineering Research Center of Network Management Technology for High-Speed Railway of Ministry of Education, Beijing Jiaotong University, Beijing, China.
\protect\\
E-mail: bogao@bjtu.edu.cn.
\IEEEcompsocthanksitem Corresponding author: Yuwei Wang.
}% <-this % stops an unwanted space
\thanks{
	%Manuscript received January xxx, xxx; revised xxx, xxx, xxx and xxx, xxx, xxx; accepted xxx, xxx, xxx. Date of publication xxx, xxx, xxx; date of current version ???, ???, ???. 
	This work was supported by the National Key Research and Development Program of China (2021YFB2900102) and the National Natural Science Foundation of China (No. 61732017, No. 62072436, No. 62002346 and No. 61872028). }
}

%\markboth{IEEE TRANSACTIONS ON PARALLEL AND DISTRIBUTED SYSTEMS, ~Vol.~XXX, No.~XXX, XXX~XXXX}%
\markboth{{{Accepted by IEEE TRANSACTIONS ON PARALLEL AND DISTRIBUTED SYSTEMS. Copyright may be transferred without notice.}}}
{Shell \MakeLowercase{\textit{et al.}}: Bare Demo of IEEEtran.cls for Computer Society Journals}




\IEEEtitleabstractindextext{%
\begin{abstract}
\justifying
The growing interest in intelligent services and privacy protection for mobile devices has given rise to the widespread application of federated learning in Multi-access Edge Computing (MEC). Diverse user behaviors call for personalized services with heterogeneous Machine Learning (ML) models on different devices. 
Federated Multi-task Learning (FMTL) is proposed to train related but personalized ML models for different devices, whereas previous works suffer from excessive communication overhead during training and neglect the model heterogeneity among devices in MEC.
Introducing knowledge distillation into FMTL can simultaneously enable efficient communication and model heterogeneity among clients, whereas existing methods rely on a public dataset, which is impractical in reality.
To tackle this dilemma, \textbf{Fed}erated Mult\textbf{I}-task Distillation for Multi-access Edge \textbf{C}ompu\textbf{T}ing (FedICT) is proposed. FedICT direct local-global knowledge aloof during bi-directional distillation processes between clients and the server, aiming to enable multi-task clients while alleviating client drift derived from divergent optimization directions of client-side local models. Specifically, FedICT includes Federated Prior Knowledge Distillation (FPKD) and Local Knowledge Adjustment (LKA). FPKD is proposed to reinforce the clients' fitting of local data by introducing prior knowledge of local data distributions. Moreover, LKA is proposed to correct the distillation loss of the server, making the transferred local knowledge better match the generalized representation. 
Extensive experiments on three datasets demonstrate that FedICT significantly outperforms all compared benchmarks in various data heterogeneous and model architecture settings, achieving improved accuracy with less than 1.2\% training communication overhead compared with FedAvg and no more than 75\% training communication round compared with FedGKT in all considered scenarios.
%To validate the effectiveness of our proposed methods, we conduct comprehensive experiments on both image classification and transportation mode detection in comparison with six benchmarks. Results demonstrate that our proposed methods significantly outperform all benchmarks on CIFAR-10, CINIC-10 and TMD datasets in various data heterogeneous and model architecture settings, requiring less than 1.2\% training communication overhead compared with FedAvg and no more than 75\% training communication round compared with FedGKT in all considered scenarios.
\end{abstract}
}
\maketitle

% - because all conference papers position the abstrac like regular
\IEEEdisplaynontitleabstractindextext

% \fi
\IEEEpeerreviewmaketitle
\IEEEraisesectionheading{\section{Introduction}}
\IEEEPARstart{M}{ulti-access} Edge Computing (MEC) pushes computation and memory resources to the network edge, enabling low communication latency and convenient services for accessed devices \cite{cruz2022edge}.
%Along with the development of wireless network technology and the proliferation of mobile devices, large amounts of decentralized data are being generated from diverse devices, and MEC shows great potential to process such distributed data \cite{kong2022edge}. 
Along with the development of wireless network technology and the proliferation of mobile devices, increasing amounts of distributed data generated in diverse devices are processed in MEC scenarios.
%, owing to reducing data transmission bandwidth and protecting device privacy \cite{kong2022edge}.
Besides, the growing interest in edge intelligence services motivates the prominent demands for deploying Machine Learning (ML) models on devices.
Whereas for privacy concerns, collecting data from devices to the remote server for model training is often prohibited \cite{tak2020federated}.
%Traditional centralized paradigms require collecting data from devices to the remote server for model training, which inevitably results in private disclosure of devices \cite{european2016regulation}.

%Ensuring data kept in devices, Federated Learning (FL) \cite{yang2019federated} opens a new horizon for training ML models in MEC.
Federated Learning (FL) \cite{yang2019federated} opens a new horizon for training ML models in a distributed manner while keeping private data locally, and is well suited for privacy-sensitive applications in MEC, such as the internet of vehicles \cite{lim2021towards,sun2022edge}, healthcare \cite{antunes2022federated,nguyen2022federated}, etc.
%Since private data from clients are not required to share, FL is well suited for privacy-sensitive applications in MEC, such as the internet of vehicles \cite{lim2021towards,sun2022edge}, healthcare \cite{antunes2022federated,nguyen2022federated}, etc.
%However, due to different user behaviors, local data distributions across devices often vary in MEC \cite{zhou2022sourcetarget}, leading to the shared model trained through most FL methods often generalize poorly to personalized clients' data \cite{tan2022towards,cho2021personalized,yu2020salvaging,kulkarni2020survey}. Therefore, training models with different decision boundaries on clients is urgently needed.
However, local data distributions across devices usually exhibit discrepant characteristics and evident skews in MEC due to diversified individual behaviors \cite{zhou2022sourcetarget}.
This phenomenon poses requirements to inconsistent update targets among client-side local models, and thus the shared server-side global model trained through conventional FL methods generalizes poorly on heterogeneous local data \cite{tan2022towards,cho2021personalized,yu2020salvaging,kulkarni2020survey}.
%However, the application of FL to MEC faces significant challenges due to limited system resources of devices, the high degree of system heterogeneity, and variations in preferred decision boundaries of models across devices. Conventional FL alls for all clients to adopt the same model and exchange large-scale model parameters during training. Nevertheless, the same model is difficult to adapt and generalize to all clients, and the communication overhead is challenging for devices.

%Federated Multi-task Learning (FMTL) \cite{smith2017federated} fits related but personalized local models by considering each client as a task in Multi-task Learning (MTL).
%To enable the server to capture generalizable representations for all clients while allowing training differentiation among local models, 
To collaboratively train separate models with different update targets, Federated Multi-task Learning (FMTL) \cite{smith2017federated} regards local model training on each device as a learning task to fit personalized requirements.
%differentiated preferred decision boundaries across local models. 
%Since FMTL enables customized models for different clients with higher User model Accuracy (UA) \cite{mills2021multi} than traditional FL, it is considered a highly promising FL paradigm for MEC. 
However, most existing FMTL methods face two challenges to tackle in MEC.
On the one hand, exchanging large-scale model parameters or gradients during training is unaffordable for devices with inferior communication capabilities \cite{sattler2021cfd,wu2022communication}.
On the other hand, personalized models with heterogeneous model architectures are required to be deployed on clients since differentiated computational capabilities, energy states and data distributions are ubiquitous among clients \cite{tak2020federated,tang2022computational,yu2021toward}.
Whereas existing FMTL methods \cite{mills2021multi,marfoq2021federated,dinh2021new,jamali2022federated} require large-scale parameters transmission as well as only support adopting the same model architecture on the server and clients, hence are unavailable when local models are heterogeneous in MEC with constrained resources.
%A more recent approach \cite{cao2022cofed} can conduct FMTL over heterogeneous models but requires additional unlabelled data, which is inaccessible for MEC in reality.

One prospective way to avoid large-scale parameters transmission and enable heterogeneous models in FMTL is to introduce Knowledge Distillation (KD) \cite{hinton2015distilling,wang2021knowledge} as an exchange protocol across model representations (called Federated Distillation, FD), transferring knowledge or intermediate features instead of model parameters between the server and clients.
%\textcolor{black}{Knowledge Distillation (KD) \cite{hinton2015distilling,wang2021knowledge} as an exchange protocol across model representations is a potential way to avoid large-scale parameter transmission and enable heterogeneous models in FMTL, where knowledge or intermediate features instead of model parameters are transferred between the server and clients (we call such methods Federated Distillation, FD).}
%FD can effectively tackle the challenge of resource constraints and heterogeneous models since it avoids exchanging large-scale model parameters during training and allows discrepant model architectures on clients.
%FD avoids exchanging large-scale model parameters during training and allows discrepant model architectures on clients.
%However, existing FD methods either fails to consider different tasks across clients \cite{itahara2021distill,zhu2021data,cheng2021fedgems,he2020group,lin2020ensemble,li2019fedmd,wu2022exploring} or require a publicly available dataset during training  \cite{cho2021personalized,zhang2021parameterized,zhou2022sourcetarget}, which is not suitable for FMTL in MEC.
%However, most existing FD methods focus only on communication efficiency \cite{itahara2021distill,zhu2021data,sattler2021cfd}, computation \cite{cheng2021fedgems,he2020group} or model heterogeneity \cite{lin2020ensemble,li2019fedmd,wu2022exploring}, where the training of all clients are treated as the same task, without considering local data distributions. 
However, all existing FD methods that support multi-task clients \cite{cho2021personalized,zhang2021parameterized} are built on frameworks that rely on public datasets whose data distribution should be close to private data on clients \cite{liu2022communication}. 
Since collected public data needs to be compared with the clients’ private data on data distributions, all FD methods rely on public datasets will undoubtedly lead to privacy leakage of clients and are impractical in MEC \cite{afonin2022towards,yu2021toward}.
%\textcolor{black}{Nonetheless, all existing FD methods that support multi-task clients \cite{cho2021personalized,zhang2021parameterized} are based on frameworks that rely on public datasets whose data distribution should be similar to that of private client data \cite{liu2022communication}. 
%Since collected public data needs to be compared with clients' private data on data distributions, all FD methods that rely on public datasets will inevitably result in clients' privacy leakage and are impractical in MEC.}
%approaches rely on access to a pulic dataset with a data distribution as close as possible to the client's private data. The process of collecting this dataset can compromise client privacy and is therefore impractical in MEC.
%However, a publicly available datasets are required by existing FD methods that support multi-task clients \cite{cho2021personalized,zhang2021parameterized,zhou2022sourcetarget}, which is not accessible in reality. 
%Though few FD approaches that do not require public datasets simultaneously enable efficient communication and model heterogeneity \cite{he2020group,wu2022exploring}, their training of all clients are treated as a single task without considering local data distributions.
Although few FD approaches can achieve client-server co-distillation without public datasets \cite{he2020group,wu2022exploring}, 
they are only appropriate to the single-task setting because of neglecting data discrepancy among clients.
%As a result, their performance in the multi-task setting is desperate to be improved.
%In addition, FD-based FMTL methods exacerbate client drift, where client-side local optimization direction inevitably deviates from that of the global model in the presence of tasks difference among clients \cite{karimireddy2020SCAFFOLD,li2021model}.
%This phenomenon causes unsatisfactory global convergence and dramatically limits the individual performance of clients.
\textcolor{black}{However, directly imposing individualized parameters update on local models in the above FD approaches without public datasets \cite{he2020group,wu2022exploring} is commonly ineffective, since it aggravates local optimization directions deviating from that of the global model, i.e., client drift, which causes unsatisfactory global convergence and severely limits the individual performance of clients in turn} \cite{cho2021personalized,zhang2021parameterized,zhou2022sourcetarget}.
\textcolor{black}{How to overcome the negative effect of client drift and achieve local distillation differentiation without public datasets becomes the major technical challenge in FD-based FMTL.}
%How to overcome the adverse impact of client drift and well achieve local distillation differentiation becomes the primary issue in FD-based FMTL without the assistance of public datasets.
%We suggest that client drift is the core of the trade-off in FD-based FMTL, but has been ignored in previous works \cite{cho2021personalized,zhang2021parameterized}. %Therefore, we propose to establish an FD-based FMTL framework without public datasets while disaffecting local and global knowledge to mitigate the impact of multi-tasking on the design trade-offs.

%In addition, client drift \cite{karimireddy2020SCAFFOLD,li2021model} is significantly exacerbated in FD-based FMTL, where client-side local optimization direction inevitably deviates from that of the global model in the presence of tasks difference among clients.
%to achieve multi-tasking on client-side local models, 
%This phenomenon causes unsatisfactory convergence on the server due to the divergence of transferred local knowledge and ulteriorly limits the training performance of clients.

%In this paper, we first propose a FD-based FMTL framework, \textbf{Fed}erated Mult\textbf{I}-task Distillation for Multi-access Edge \textbf{C}ompu\textbf{T}ing (FedICT), for bridging FMTL with MEC. 
In this paper, we propose an FD-based FMTL framework for MEC without a public dataset, named \textbf{Fed}erated Mult\textbf{I}-task Distillation for Multi-access Edge \textbf{C}ompu\textbf{T}ing (FedICT).
%FedICT leverages KD as a flexible representation transfer protocol in substitution of model parameter exchange, thus allowing to conduct FMTL over heterogeneous models with low communication overhead.
%Different from previous FD methods that do not require a public dataset  \cite{he2020group,wu2022exploring}, 
FedICT enables differentiated learning on client-side local models via distillation-based personalized optimization while disaffecting the knowledge transferred between the server and clients, so as to mitigate the impact of client drift on model convergence while enabling personalized local models.
%to simultaneously achieve generalized global model and personalized local models.
%to mitigate the impact of multi-task clients on the design trade-offs.
%to keep the server-side global model representations universal to all clients, while 
Specifically, FedICT consists of two parts, Federated Prior Knowledge Distillation (FPKD) for personalizing client-side distillation and Local Knowledge Adjustment (LKA) for correcting server-side distillation.
The former enhances clients' multi-task capability based on prior knowledge of local data distributions and reinforces the fitting degree of local models to their local data by controlling class attention during local distillation.
%through adaptively controlling the class focus during local distillation.
%based on class distributions during local distillation.
%To prevent poor convergence derived from inconsistent structural information between local and global knowledge, 
The latter is proposed to correct the loss of global distillation on the server, which prevents the global optimization direction from being skewed by local updates.
%which can prevent global optimization direction drift away from local optimization directions under deflective data distributions.
To our best knowledge, \textbf{this paper is the first work to investigate federated multi-task distillation without additional public datasets in multi-access edge computing,} which realizes multi-task training requirements in a communication-efficient and model-heterogeneity-allowable manner, and is practical for MEC.

In general, our contributions can be summarized as follows:
\begin{itemize}
	\item
    We propose a novel FD-based FMTL framework in MEC (namely FedICT), which can realize distillation-based personalized optimization on clients while reducing the impact of client drift from a novel perspective of alienating local-global knowledge without public datasets.
	%We propose a novel FD-based FMTL framework in MEC (namely FedICT), which can realize distillation-based personalized optimization on clients from a novel perspective of alienating local-global knowledge. FedICT enables the server to capture generalizable representations from corrected local knowledge and makes local models perform well over discrepant data distributions with the assistance of customized global knowledge. 
	%We propose a novel FD-based FMTL framework, FedICT, to bridge FMTL with MEC. Through combining distillation-based personalized optimization on clients with local-global knowledge aloof, the server is able to capture generalized representations from corrected local knowledge while clients can perform well on different tasks with customized global knowledge transferred.
	\item
	We propose FPKD to enhance fitting degrees of client-side local models on discrepant data via introducing prior knowledge of local data distributions.
	Further, LKA is proposed to correct the distillation loss of the server-side global model, aiming to alleviate client drift derived from knowledge mismatch between clients and the server.
	%We propose two techniques to achieve local-global knowledge aloof: FPKD, which enhances the clients' fitting degree to their respective local data by introducing the prior knowledge of local data distributions. LKA, which corrects the distillation loss on the server to better fit overall data.
	\item
	We conduct extensive experiments on CIFAR-10, CINIC-10 and TMD datasets. Results show that our proposed FedICT can improve average User model Accuracy (UA) \cite{mills2021multi} of all compared benchmarks. Besides, FedICT enables efficient communication and faster convergence, achieving the same average UA with less than 1.2\% of training communication overhead compared with FedAvg and no more than 75\% of communication rounds compared with FedGKT in all experimental settings.
 %Besides, the communication overhead required for FedICT to achieve comparable performance is less than 1.2\% of that of FedAvg. 
    %FedICT achieves faster convergence, reducing the number of communication rounds to no more than 75\% to reach the same average UA compared with FedGKT.
	%achieve significantly higher communication efficiency than conventional FL, with at least $82\times$ more efficient communication compared with FedAvg, pFedMe, FedAdam and MTFL. (3) increase the convergence speed compared with other FD methods, with more than $33\%$ faster convergence than FedGKT.
\end{itemize}

\section{Related Work}
\label{related-work}
\subsection{Federated Multi-task Learning}
%FMTL \cite{smith2017federated} is proposed to implement FL when the preferred decision boundaries across clients are differ. 
%By treating each client as a task in MTL, FMTL enables all clients to collaboratively explore a shared generalized representation while allowing personalized objectives on local models.
FMTL \cite{smith2017federated} is proposed to fit related but personalized models over FL, which enables clients to collaboratively explore a shared generalized representation while allowing personalized objectives on local models. 
%Motivated by this idea, a series of approaches are proposed that allow clients to introduce non-federated network layers \cite{mills2021multi}, adopt improved optimization objectives \cite{dinh2021new,mortaheb2022fedgradnorm}, or fit different latent distributions of clients with multiple shared component models \cite{marfoq2021federated}. 
Motivated by this idea, a series of approaches are proposed, such as introducing non-federated network layers \cite{mills2021multi}, adopting diversified optimization objectives \cite{dinh2021new,mortaheb2022fedgradnorm}, 
or leveraging ensemble models to fit client-side data distributions \cite{marfoq2021federated}.
%or fitting different latent distributions of clients with multiple shared component models \cite{marfoq2021federated}. 
Specifically, \cite{mills2021multi} allows clients to separately optimize personalization layers. 
\cite{marfoq2021federated} adopt linear combinations of multiple shared component models, assuming that data distributions of clients are a mixture of multiple unknown underlying distributions. 
%Specifically, existing approaches propose to allow clients to adopt separately optimized personalization layers \cite{mills2021multi} or use an ensemble of models to fit multiple latent distributions in a single client \cite{marfoq2021federated}. 
Some approaches utilize Laplacian Regularization to constrain local models \cite{dinh2021new} or adopt dynamic weights on local model gradients \cite{mortaheb2022fedgradnorm}. 
\textcolor{black}{Another common type of FMTL is cluster-based FL \cite{sattler2020clustered,jamali2022federated}, where clients are clustered according to data similarity andthe clients in each cluster learn a shared model.} 
\textcolor{black}{However, all the above methods adopt the traditional communication protocol represented by FedAvg \cite{mcmahan2017communication}, which requires exchanging large-scale model parameters with the same model architecture between the server and clients.}
%Notably, \textcolor{black}{Cao et al. \cite{cao2022cofed} propose a co-training-based \cite{blum1998combining} FMTL method, which allows training heterogeneous clients through fusing the views of data from each client in a semi-supervised setting.} However, this method relies on a publicly available dataset, which is impractical in reality \cite{afonin2022towards}. 
%As a result, existing FMTL methods are not practical for MEC, where each client individually designs its local model with low communication consumption.

\subsection{Federated Learning in Multi-access Edge Computing}
FL performs collaborative model training on distributed devices at the network termination, whereas these devices often possess heterogeneous system configurations and training goals with constrained resources \cite{tang2022computational,tak2020federated}. 
A series of approaches are proposed to reduce the computational or communication on devices through transferring computation burden from devices to the edge server \cite{jiang2022fedsyl}, adopting model pruning methods to lighten model sizes on devices \cite{jiang2022model}, or establishing computing- and communication-friendly training paradigm \cite{he2020group}.
%A series of approaches are proposed to address the problem of constrained computational or communication overheads of devices during FL training, through introducing the resources from the edge server in assistance of the training process on devices \cite{jiang2022fedsyl}, or adopting model pruning to lighten the size of the models at the network termination \cite{jiang2022model}.
Another line of research is to fit different requirements among devices: adopting adaptive learning rates to fit the personalized accuracy goals of clients \cite{jiang2020customized}, transferring historical information from previous personalized models to maintain local models' well performance on individual clients \cite{jin2022personalized}, or leveraging memory-efficient source-free unsupervised domain adaptation to make local models adapt their respective data \cite{zhou2022sourcetarget}.
However, none of the above approaches can simultaneously meet communication constraints and enable model heterogeneity among clients, which is inapplicable to MEC scenarios in practice.
%They cannot be used in conjunction with each other as well, hence are not practical for MEC in reality.
%Moreover, the risk of exposing the privacy of the user model architecture due to the requirement of client-global model homogeneity cannot be resolved.

\subsection{Knowledge Distillation in Federated Learning}
KD enables knowledge to be transferred from one ML model to another to facilitate constructive optimization of the latter model. 
KD has been utilized in various fields up to date, such as model compression \cite{hinton2015distilling,liu2022multi}, domain adaptation \cite{wu2021spirit,nguyen2021unsupervised,wu2021spirit2} and distributed training \cite{anil2018large,bistritz2020distributed}.
%knowledge transferred from bulky or ensemble models to light-weight models for model compression \cite{hinton2015distilling,hao2021data,liu2022multi}, knowledge transferred from one domain to another for domain adaptation \cite{wu2021spirit,nguyen2021unsupervised,wu2021spirit2}, knowledge inter-changed across models for distributed neural network training \cite{anil2018large,bistritz2020distributed}.
Jeong et al. \cite{jeong2018communication} first introduce KD to FL as an exchange protocol for cross-clients model representations, and such distillation-based FL methods are called federated distillation (FD).
%%Driven by the idea of FD, a range of algorithms leverage publicly available datasets \cite{lin2020ensemble,li2019fedmd,cheng2021fedgems,itahara2021distill,cho2021personalized,cho2022heterogeneous,zhang2021parameterized} or embedded features \cite{he2020group,wu2022exploring,chen2022metafed} to extract knowledge, and then perform client-server collaborative distillation without exchanging model parameters during the whole training process.

%FD is first proposed for FL under local models with different architectures \cite{li2019fedmd}: the server iteratively generates consensus based on the logits uploaded by clients, and then distributes them to clients for local distillation, thus exchanging information across clients in a model agnostic manner \cite{afonin2022towards}.
One of the most representative FD methods is proposed in \cite{li2019fedmd}, where the server iteratively generates consensus based on client logits and then distributes consensus to clients for local distillation.
Subsequent approaches are improved in terms of data dependency \cite{itahara2021distillation,lin2020ensemble}, knowledge distribution \cite{itahara2021distillation,chang2019cronus}, knowledge filtering or weighting \cite{cho2021personalized,zhang2021parameterized,cho2022heterogeneous,cheng2021fedgems}, etc.
Several works \cite{itahara2021distillation,lin2020ensemble} extend conventional supervised FD methods to semi-supervised paradigms.
Besides, some approaches adjust the knowledge distribution during distillation to accelerate client-side convergence \cite{itahara2021distillation} or counteract poisoning attacks \cite{chang2019cronus}.
More recent works are proposed to filter, weight, or cluster knowledge from clients with similar local data distributions \cite{cho2021personalized,zhang2021parameterized,cho2022heterogeneous,cheng2021fedgems}.
However, all the above approaches rely on public datasets whose data distribution should be similar to local training data \cite{liu2022communication}, but such datasets are hard to access in reality \cite{afonin2022towards,yu2021toward}.
%with similar distributions to clients' private data \cite{afonin2022towards}, but such datasets are not accessible in reality.
Although few approaches can realize FD without public datasets \cite{he2020group,wu2022exploring,zhu2021data,zhang2022feddtg}, they either neglect knowledge deviation of local models derived in multi-task setting \cite{he2020group,wu2022exploring}, or confront with tremendous communication overhead for exchanging model parameters \cite{zhu2021data,zhang2022feddtg}.
Therefore, existing FD methods are not suitable for FMTL in MEC.
%different preferred decision boundaries among local models \cite{he2020group,wu2022exploring}, or the large communication overhead due to the inability to avoid model parameter exchange during training \cite{zhu2021data,zhang2022feddtg}. 

\section{Notations and Preliminary}
\label{background-and-notation}
\subsection{Formulation of Federated Multi-task Learning}
This paper investigates the cross-device FMTL in which heterogeneous clients jointly train ML models coordinated by the server, with the goal of training personalized local models that can adapt to local data distributions. 
The main notations in this paper are summarized in TABLE \ref{notation}.
Without loss of generality, we study $C$ class classification in FMTL. Assuming that $K$ clients participate in FL training and $\mathcal{K}:=\{1,2,...... ,K\}$.
%this paper studies $C$ class classification in FMTL. 
Each client $k \in \mathcal{K}$ possesses a local dataset ${\hat \mathcal{D}}^k := \bigcup\limits_{i = 1}^{{N^k}} {\{ ({{\hat X}_i^k},{{\hat y}_i^k})\} }$ with $N^k$ samples.% and $C$ classes to classify.
The local dataset ${\hat \mathcal{D}}^k$ is sampled from the local data distribution ${{\cal D}^k} := \bigcup\limits_{i = 1}^\infty {\{ (X_i^k,y_i^k)\} }$, where ${{\hat {\cal D}}^k} \subset {{\cal D}^k}$.
Different from the optimization objectives of conventional FL methods \cite{mcmahan2017communication,li2020federated,reddi2021adaptive} where all clients share the same model, we expect that client $k$ obtains a local model ${{\cal F}^k(\cdot)}$ that can maximize the localized evaluation metric $\mathcal{M}(\cdot)$ for its personalized local data, i.e.,
\begin{equation}
	\mathop {\arg \max }\limits_{{W^k}} \mathop E\limits_{(X_i^k,y_i^k)\sim{{\cal D}^k}} [{\cal M}({{\cal F}^k}(X_i^k;{W^k}),y_i^k)],
\end{equation}
where $W^k$ is the parameter of the local model at client $k$. 
%Generally, FMTL guides local models to form associations with common representations of all clients during the training process, so as to improve local models' generalization performance.
Generally, FMTL guides local models to accommodate universal representations integrated from all clients during the training process, so as to improve local models' performance on local data.

\begin{table}[!t]
	\centering
	\setlength\extrarowheight{0.5pt}
	\caption{Main notations and descriptions.}
	\begin{tabular}{c|c}
		\hline
		\textbf{Notation} & \textbf{Description} \\ \hline
		$K$        & Number of clients               \\
        $R$     & Maximum number of communication rounds \\
		$\hat \mathcal{D}^k$      & Local dataset of client $k$     \\
		$N^k$      & Number of samples in $\hat \mathcal{D}^k$     \\
		${\hat X}_i^k$      & The $i$-th sample of $\hat \mathcal{D}^k$     \\
		${\hat y}_i^k$      & The label of ${\hat X}_i^k$     \\
		$W^S$ & The global model parameters of the server \\
		$W^k$ & The local model parameters of client $k$ \\
		%$W^k_e$ & The feature extractor parameters of client $k$\\
		%$W^k_p$ & The predictor parameters of client $k$\\
		$z_{{\hat X}_i^k}^S$ & The global knowledge of ${\hat X}_i^k$\\
		$z^k_{{\hat X}_i^k}$ & The local knowledge of ${\hat X}_i^k$\\
		${{\hat H}_i^k}$ & The extracted features of ${\hat X}_i^k$\\
		${{d}^k}$ & The local data distribution vector of client $k$\\
		${{d}^S}$ & The global data distribution vector\\
		$J_{ICT}^S$ & \begin{tabular}[c]{@{}c@{}}The optimization objective of global model \\when adopting FedICT\end{tabular}
		\\
		$J_{ICT}^k$ & \begin{tabular}[c]{@{}c@{}}The optimization objective of local model \\on client $k$ when adopting FedICT\end{tabular}
		\\
		%		%$u_i^k$ & The $i^{th}$ dimension of the vector $z^k_{X^k}$\\
		%		%$v_i^k$ & The $i^{th}$ dimension of the softmax output of $z^k_{X^k}$\\
		%		%$\varepsilon$ & Upper bound on tolerable distribution difference\\
		%		$\theta$ & The parameter of the auxiliary mapping in $\psi(\theta;\cdot)$\\
		%		$m$ & The index of the maximum element in $p_{X^k}^k$ \\
		%		$t$ & The input scaling parameter of kernel function \\
		%		$T$ & The hyper-parameter of target peak probability\\	
		%		$E$ & The hyper-parameter of target Shannon-entropy  \\
		%		$\tau(\cdot)$ & The softmax mapping \\
		%		%$f(W^*;\cdot)$ & Nonlinear function determined by the weights $W^*$ \\
		%		%$arch(\cdot)$ & Neural network architecture function \\
		%		%$acc(\cdot)$ & Precise metric function \\
		%		$L_{CE}(\cdot)$ & The cross-entropy loss function \\
		%		$L_{sim}(\cdot)$ & The knowledge-similarity loss function \\
		%		%$L_{S}(\cdot)$ & The loss function on the server \\
		%		%$KL(\cdot)$ & The KL divergence loss function \\
		%		$\max(\cdot)$ & The maximum function \\
		%		$\varphi(\cdot)$ & The refinement mapping over distributed knowledge \\
		%		%$dist(\cdot)$ & Metric of knowledge distribution among clients\\
		%		%$L_{DKC}(\cdot)$ & The DKC loss \\
		%		%$L_{S}^{'}(\cdot)$ & Distillation loss function on server with DKC \\
		%		$\sigma ( \cdot )$ & The kernel function in KKR \\
		%		%$\max(\cdot)$ & The maximum input probabilities \\
		%		%$\varphi_{KKR}(\cdot)$ & The KN-DKCM \\
		%		$H(\cdot)$& The Shannon-entropy function \\
		%		$\psi(\cdot)$ & The auxiliary mapping in SKR\\
		%		%$\varphi_{SKR}(\cdot)$ & The SH-DKCM \\
		%		%此表格仅展示了preliminary中的符号
		\hline
	\end{tabular}
	\label{notation}
\end{table}

\begin{figure*}[t!]
	\centering
	\includegraphics[width=1.0 \textwidth]{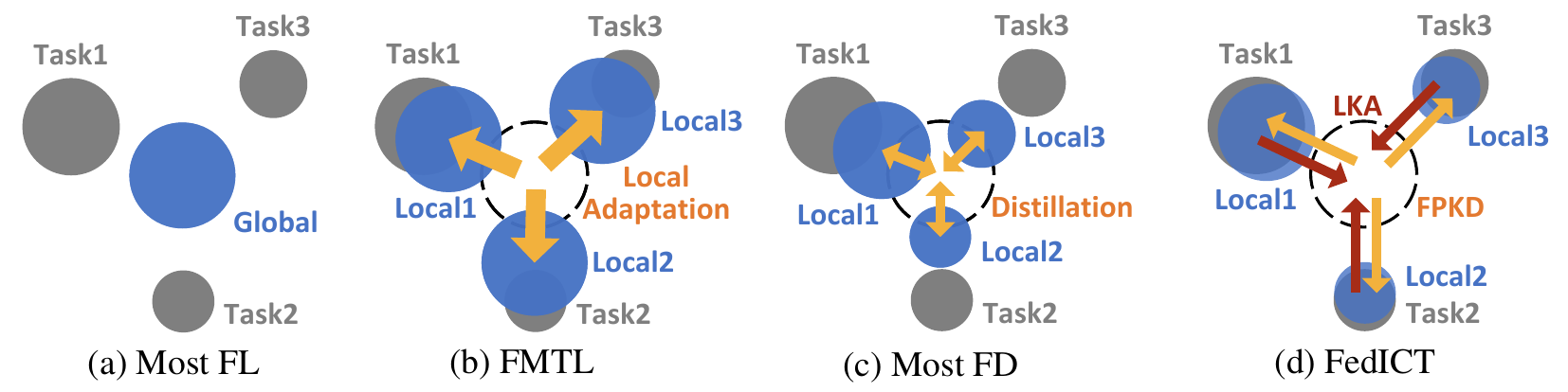}
	\caption{Comparison of different FL methods in MEC. Grey circles indicate the parameter requirements for different training tasks on devices, and the blue circles indicate the trained model parameters. Each circle's size represents the scale of model parameters, and the distance between two arbitrary circles implies the degree of differences between their corresponding parameters.} %caption是图片的标题
	\label{local-global} 
\end{figure*}

\subsection{Basic Process of Federated Distillation}
This paper follows the framework of proxy-data-free FD \cite{he2020group,wu2022exploring}, 
where the model of arbitrary client $k$ is divided into two parts, the feature extractor and the predictor with corresponding parameters $W^k_e$ and $W_p^k$ respectively. 
Hence, the model parameters of client $k$ are denoted as $W^k:=\{W^k_e,W^k_p\}$. 
The server adopts a global model with only the predictor to synthesize local knowledge, whose parameters are denoted as $W^S$. 
It is worth noting that the inputs of all feature extractors and the outputs of all predictors share the same shape.
%the output shapes of feature extractors for all local models are the same, \textcolor{black}{and their predictors accept the same size of inputs as the global model.}

Proxy-data-free FD relaxes the requirements of model homogeneity and decreases the communication overhead through exchanging knowledge or features in replacement of model parameters between the server and clients.
The overall training procedure consists of multiple communication rounds, and each round adopts a stage-wise training paradigm, successively updating global and local model parameters in a co-distillation manner \cite{anil2018large}.
Specifically, let $f(\cdot;W^*)$ denotes the non-linear mapping determined by the parameters ${W^ * } \in \{ \bigcup\limits_{k = 1}^K {{W^k}}  \cup {W^S}\} $, and $R$ denotes the maximum number of communication rounds. $\tau(\cdot)$ is the softmax mapping, $L_{CE}(\cdot)$ is the cross-entropy loss function, and $L_{sim}(\cdot)$ is the customized knowledge similarity loss function, which takes KL divergence loss by default. Throughout the training process, we refer to the logits from clients as local knowledge and the logits from the server as global knowledge.

The basic process of FD can be divided into two stages as follows:
\begin{itemize}
	\item
	\textbf{Local Distillation.} Client $k$ updates its local model parameters $W^k$ based on the local labels ${\hat y}_i^k$ and the downloaded global knowledge $z_{\hat X_i^k}^S$. The basic objective of local model optimization on client $k$ $J^k(\cdot)$ can be expressed as follows:
	\begin{equation}
		\begin{array}{l}
			 \; \; \; \; \arg \mathop {\min }\limits_{{W^k}} {J^k}({W^k})\\
			 = \arg \min \limits_{{W^k}} \mathop E\limits_{(\hat X_i^k,\hat y_i^k)\sim {\hat \mathcal{D}}^k} [{L_{CE}}(\tau (f(\hat X_i^k;{W^k})),\hat y_i^k)\\
			 \; \; \; \; + \beta \cdot{L_{sim}}(\tau (f(\hat X_i^k;{W^k})),\tau(z_{\hat X_i^k}^S))],\\
		\end{array}
	\label{local-disill-base}
	\end{equation}
	where $z_{\hat X_i^k}^S$ is the global knowledge extracted from the local features $\hat H_i^k$ in the previous communication round, which is derived by:
	\begin{equation}
		z_{\hat X_i^k}^S = f(\hat H_i^k;{W^S}).
		\label{z-S}
	\end{equation}
	\item
	\textbf{Global Distillation.} The server updates the global model parameters $W^S$ based on the uploaded local knowledge $z_{\hat X_i^k}^k$, the uploaded local features $\hat H_i^k$ and labels ${\hat y}_i^k$. The basic objective of global model optimization $J^S(\cdot)$ can be expressed as follows:
	\begin{equation}
		\begin{array}{l}
			\; \; \; \; \arg \min \limits_{{W^S}} J^S({W^S})\\
			= \mathop {\arg \min }\limits_{{W^S}} \mathop {{\rm{ }}E}\limits_{(\hat X_i^k,\hat y_i^k)\sim\bigcup\limits_{k \in {\cal K}} {{{\hat {\cal D}}^k}} } [{L_{CE}}(\tau (f(\hat H_i^k;{W^S})),\hat y_i^k)\\
			\; \; \; \; + \beta  \cdot {L_{sim}}(\tau (f(\hat H_i^k;{W^S})),\tau(z_{\hat X_i^k}^k))],\\
		\end{array}
	\label{global-distill-base}
	\end{equation}
	where $\hat H_i^k$ and $z_{\hat X_i^k}^k$ are the local features and knowledge of client $k$ generated in the last local distillation process.
	They can be derived by:
	\begin{equation}
		\hat H_i^k = f(\hat X_i^k;W_e^k),
		\label{H-k}
	\end{equation}
	\begin{equation}
		z_{\hat X_i^k}^k = f(\hat X_i^k;{W^k}).
		\label{z-k}
	\end{equation}
\end{itemize}
Local and global distillation stages are alternately executed until model convergence.
As only embedded features, logits, and labels are exchanged between the server and clients and their sizes are much smaller than model parameters \cite{wu2022exploring,he2020group}, FD can naturally guarantee communication effectiveness.
Furthermore, FD does not require homogeneous model architectures on clients and thus can support various devices with different system configurations.

\begin{table*}[]
	\setlength\extrarowheight{-1.0pt}
	\renewcommand\arraystretch{1.2}
	\centering
	%\caption{Comparison of FedICT with other FL methods. Methods that satisfy all four given conditions are considered to be deployed practically in MEC.}
	\caption{\textcolor{black}{Comparison of FedICT with other FL methods in terms of four indicators that characterize whether FL method is practically deployable in MEC.}}
	\begin{tabular}{l|c|c|c|c}
		\hline
		$\quad \quad \quad \quad \; \; \; \;\; \; \; \; \; \; \; \;$ \textbf{Method}& \begin{tabular}[c]{@{}c@{}}\textbf{Task Hetero.}\\ \textbf{Among Clients}\end{tabular} & \begin{tabular}[c]{@{}c@{}}\textbf{Model Hetero.}\\ \textbf{Among Clients}\end{tabular} & \begin{tabular}[c]{@{}c@{}}\textbf{Efficient}\\ \textbf{Communication}\end{tabular} & \begin{tabular}[c]{@{}c@{}}\textbf{Do Not Require} \\ \textbf{Public Data}\end{tabular} \\ \hline
		FedAvg \cite{mcmahan2017communication} /FedProx \cite{li2020federated}/FedAdam \cite{reddi2021adaptive}  & \XSolidBrush                                                                          & \XSolidBrush                                                                           & \XSolidBrush                       & \Checkmark                                                                                \\
		pFedMe \cite{t2020personalized}/FedEM \cite{marfoq2021federated}/MTFL \cite{mills2021multi}  & \Checkmark                                                                          & \Checkmark                                                                           & \XSolidBrush                       & \Checkmark                                                                                \\
		FedMD \cite{li2019fedmd}/DS-FL \cite{itahara2021distillation}/FedGEMS \cite{cheng2021fedgems}      & 
		\XSolidBrush
		& \Checkmark                                                                           & \Checkmark                       & \XSolidBrush                                                                                \\
		PERFED-CKT \cite{cho2021personalized}/KT-pFL \cite{zhang2021parameterized}/CoFED \cite{cao2022cofed}& \Checkmark                                                                          & \Checkmark                                                                           & \Checkmark                       & \XSolidBrush                                                                                \\
		FedGKT \cite{he2020group}/FedDKC \cite{wu2022exploring}          & \XSolidBrush                                                                         & \Checkmark                                                                           & \Checkmark                       & \Checkmark                                                                                \\ \hline
		\textbf{FedICT}                  & \Checkmark                                                                          & \Checkmark                                                                           & \Checkmark                       & \Checkmark                                                                                \\ \hline
	\end{tabular}
	\label{cmp-table}
\end{table*}

\section{Federated Multi-task Distillation for Multi-access Edge Computing}
\label{framework}
\label{methods}
\subsection{Motivation}
\subsubsection{Superiority of FD for FMTL in MEC}
%\subsubsection{Why FD for FMTL in MEC?}
The core challenges of FMTL in MEC are twofold: limited communication capabilities and heterogeneous models.
\begin{itemize}
	\item
	\textbf{Limited Communication Capabilities.} Devices possess poor communication capabilities and are unable to communicate at scale \cite{tang2022computational,tak2020federated,sattler2021cfd,wu2022communication}.
    %\textcolor{blue}{Device communication capabilities are commonly deficient due to low hardware configurations and strict power control \cite{tang2022computational,tak2020federated}.}
	%devices possess poor communication capabilities \cite{tang2022computational,tak2020federated} and are unable to communicate at scale \cite{he2020group,cheng2021fedgems,zhou2022sourcetarget}.
	\item
	\textbf{Heterogeneous Models.} Each client call for independently designed models with differentiated parameters to satisfy personalized requirements since devices vary in computational capabilities, energy states and data distributions \cite{tak2020federated,tang2022computational,yu2021toward}.
	%Clients' forms, manufacturers, and device capabilities vary widely \cite{tak2020federated}. 
\end{itemize}
Most FMTL methods require to exchange large-scale model parameters during training. Hence, tremendous communication overhead is a key trouble when deploying to MEC. 
In addition, model heterogeneity combined with multi-tasking is also a big issue in MEC, as shown in Fig \ref{local-global}. 
As displayed in Fig. \ref{local-global} (b), although existing FMTL methods can capture common representations between interrelated tasks and generalize well to different tasks via local adaptation, they fail to deploy models with suitable parameters size for each client.
%Furthermore, "client drift" is easy to occur in FMTL where client-side local optimization direction deviates from that of the global model. Specifically, the server converges slowly due to aggregation of client parameters that drift away from the global model, which in turn degrade the performance of the overall FL system \cite{karimireddy2020SCAFFOLD,li2021model}.
%However, as tasks on clients differ, end-to-end FMTL inevitably suffers from discrepancies in clients' optimization direction. Suppressing such discrepancies runs counter to the need for clients to perform local tasks. The above dilemma poses severe trouble for FMTL methods based on parameter aggregation.

We claim that adopting FD for FMTL in MEC has the following advantages:
\begin{itemize}
	\item
	\textbf{Communication Efficiency.} 
	The size of knowledge or embedded features exchanged between the server and clients are much smaller than that of model parameters. As a result, FD-based FMTL methods are effective in MEC scenario, where communication resources among clients are strictly limited.
	\item
	\textbf{Heterogeneous Models Supportability.} 
	Even if clients adopt independent models with various architectures, FD-based FMTL can be deployed and trained as long as few preconditions are met (e.g. agreement on the size of knowledge or features), which is applicable to MEC.
	\item
	\textbf{Multi-task Feasibility.}
	Local distillation can be tailored to adapt local data distributions, meeting client-side local task requirements.
	%In FD-based FMTL, the distillation process on global knowledge can be tailored to adapt local client data distributions, allowing for the satisfaction of various client-side local task requirements.
	%conforming to clients' need to perform well on respective local tasks. 
	%At the same time, the transferred structured information of local knowledge can be customized to correct for the global representation from biased optimized. As the global model does not require directly fitting differentiated local clients, the FD system is expected to converge stably.
\end{itemize}

In general, adopting FD for FMTL is a feasible choice for MEC: it not only meets the communication limitation and model heterogeneity requirements of MEC, but also enables collaborative training among clients with different tasks.

%\subsubsection{Why Local-global Knowledge Aloof in FD?}
\subsubsection{Insight of Aloof Local-Global Knowledge in FD}
Since FD requires local models to mimic the global model partially, local models tend to learn an isomorphic representation of the global model, somewhat inhibiting the ability to accommodate multiple tasks on clients.
%somewhat inhibiting their ability to perform well on multiple tasks. 
As shown in Fig. \ref{local-global} (c), all clients tend to learn a common representation that is similar to the server in existing FD methods, and fail to perform well on different local tasks due to ignoring adapt local models to local data \cite{he2020group,wu2022exploring}.
%as they ignore different data environments on diverse clients \cite{he2020group,wu2022exploring}.
Furthermore, as FMTL expects to train local models with a high degree of personalization, it raises a question of how the global model learns a uniform generalizable representation from highly biased local knowledge: local models need to perform well on heterogeneous local data distributions, and their inductive preferences necessarily deviate from that of the global model, which in turn increases the difficulty of distillation-based fusion of local knowledge.

Based on the above analysis, we suggest that \textbf{knowledge correction is necessary during local and global distillation. Therefore, we expect to inject localized prior knowledge in local distillation and de-localize local knowledge in global distillation, i.e., keeping local-global knowledge aloof.} 
Based on the customized local distillation objective, each local model can better adapt to the local task. Based on the de-localized global distillation objective, the global model can converge stably towards global generalization. 
Through adopting this idea, the server can learn generalizable knowledge while clients possess satisfactory capabilities of learning discrepant local tasks, with different representations between the server and clients.
%Posterior to this idea, with vastly different server and client representations and preferred decision boundaries, the server can still learn generalizable knowledge while the clients maintain strong local task capability.

Based on the above insight, FedICT is proposed, whose optimization sketch map in MEC is shown in Fig. \ref{local-global} (d), and comparisons with other FL methods are listed in TABLE \ref{cmp-table}. 
Compared with the state-of-the-art methods, our proposed FedICT not only allows task and model heterogeneity among clients, but also enables efficient communication without the assistance of a public dataset, which can be deemed as the first FD work on multi-task setting to be practically deployed in MEC.
%which is the first FD work that is actually deployable in MEC.

\subsection{Framework Formulation}
Different from previous methods \cite{he2020group,wu2022exploring}, 
%we perform a knowledge adaptation process for knowledge transferred from both the server and the clients: adapting knowledge extracted from the locally distributed data to overall data in the global distillation process, and introducing prior knowledge from the local data distributions during the local distillation process, as shown in Fig. \ref{main}.
we perform knowledge adaptation processes in both local and global distillation stages.
%\textcolor{black}{Specifically, clients train personalized local models based on the prior knowledge of local data distributions during local distillation, and the server aggregates uploaded local knowledge according to the global-local deviation during global distillation.}
Specifically, prior knowledge of local data distributions is introduced to personalize local models during local distillation; the discordance of global versus local data distributions is considered to reduce global-local knowledge divergence during global distillation.
%through inducing local knowledge to match the overall data. 
% \color{red}
% Fig. \ref{main} displays the framework of our proposed FedICT, where the red and blue components represent the hardware facilities, data information, knowledge, or processing modules on the edge server and the network termination, respectively. As shown, smoothed knowledge transferred from the server (red solid line) is personalized (red dashed line) based on the skewed local data distribution information (blue bars) to adapt to local tasks. Biased knowledge transferred from clients (solid blue line) is corrected (blue dashed line) via simultaneous considering relatively balanced global and skewed local data distribution information (red and blue bars).
% \color{black}

To be specific, we define $d^k:=dist({\hat {\cal D}^k})$ as the local data distribution vector of client $k$ and $d^S:=dist(\bigcup\limits_{k = 1}^K {{\hat {\cal D}^k}})$ as the global data distribution vector, where $dist(\cdot)$ maps the input dataset to its corresponding data distribution vector for estimating the data distribution of a given dataset.
In this paper, we adopt data category frequencies tepresent data distributions. For any dataset ${{\hat \mathcal{D}}^*}: = \bigcup\limits_{i = 1}^{N^*} {\{ (\hat X_i^*,\hat y_i^*)\} }$ with $N^*$ samples, the $i$-th dimension of its data distribution vector $dist{({\hat \mathcal{D}^*})_i}$ depends on the frequency of its $i$-th class $f_i^*$, that is:
\begin{equation}
	dist{({{\hat \mathcal{D}}^*})_i} =f_i^*= \frac{{\sum\limits_{{y_i^*} \in {\mathcal{D}^*}} {\delta ({y_i^*} = i)} }}{{{N^*}}},
	\label{dist-k}
\end{equation}
where $\delta(\cdot)$ is an indicator function that returns 1 when the input equation holds and 0 otherwise.

During local distillation, local models are updated with reference to local data distribution information, aiming to achieve superior performance on local tasks.
Specifically, we formulate the new local distillation objective $J_{ICT}^k(\cdot)$ for client $k$ as follows:
\begin{equation}
	\begin{array}{l}
		\; \; \; \; \arg \min \limits_{{W^k}} J_{ICT}^k({W^k})\\
		=\arg \min \limits_{{W^k}} [J^k({W^k})+\lambda  \cdot J^k_{FPKD}(W^k;d^k)],
	\end{array}
	\label{clinet-lga-gener}
\end{equation}
where $J^k_{FPKD}(\cdot)$ is the optimization component of client $k$ based on the distribution vector of local data $d^k$. 
%We expect that each client $k$ learns a better local model based on the prior knowledge of corresponding local data distribution $d^k$.
%We expect that each client $k$ learns better local model decision boundary as indicated by the prior knowledge of corresponding local data distribution $d^k$, and thus adapts more optimally to future local data.

%During local distillation, local models are updated with reference to local data distribution information, aiming to achieve superior performance on local tasks.
During global distillation, the global model is updated considering the discordance of the global versus local data distributions, realizing the global knowledge de-localization to maintain the global model's global-to-local perspective rather than a narrow local perspective.
%\textcolor{black}{During global distillation, we expect the global model to maintain the global-to-local perspective, rather than only the local perspective.}
%Thus, the discordance of the global versus local data distributions should be considered in the global optimization objective to realize the global model's de-localization.
%Therefore, the discordance of the global versus local data distributions needs to be taken into account as the optimizing component for global model's de-localization. 
Specifically, we formulate the new global distillation objective $J^S_{ICT}({W^S})$ as follows:
\begin{equation}
	\begin{array}{l}
		\; \; \; \; \arg \min \limits_{{W^S}} J^S_{ICT}({W^S})\\
		=\arg \mathop {\min }\limits_{{W^S}} [ J^S({W^S}) + \mu  \cdot J^S_{LKA}(W^S;d^S,d^k)],
	\end{array}
	\label{server-lga-gener}
\end{equation}
where $J^S_{LKA}(\cdot)$ is the optimization component based on the de-localized local knowledge. 
%We expect that the server maintains a more stable global perspective guided by the corrected local knowledge, and enables to highly tolerate task heterogeneity of clients.
%enables to learn from local knowledge with a high degree of tolerance for task heterogeneous clients.

%In general, local distillation process introduces the prior knowledge of local data distributions to obtain preferences of local models, enabling to better suit the local tasks.
%global distillation process corrects local knowledge to better match the overall data, enabling the server to capture generalized representations from differentiated local knowledge to improve client-side performance when the global knowledge transfers back.
%Throughout the whole FD process, the representations of both global and local models are biased to the distribution of their respective related data, i.e. presenting aloof.
In general, we anticipate that the transferred knowledge from both global and local models will be biased toward the data distribution associated with their respective target models, i.e. inducing aloof local-global knowledge.
Such induction during bi-directional distillation processes enables local models to sufficiently fit local tasks while facilitating the global model to integrate personalized local knowledge for achieving faster convergence.
Specifically, we propose Federated Prior Knowledge Distillation (FPKD, related to $J^k_{FPKD}$) and Local Knowledge Adjustment (LKA, related to $J^S_{LKA}$) to jointly achieve aloof local-global knowledge.
The details of our proposed techniques are described in the following sections.

% \begin{figure*}[t!]
% 	\centering
% 	\includegraphics[width=1.0 \textwidth]{main.pdf}
% 	\caption{Framework of FedICT. (1) Transmit stream. (2) Upload/Download stream. (3) Neural network modules. (4) Knowledge/Data distribution vector processing. (5) Data distribution. (6) Knowledge distribution.} %caption是图片的标题
% 	\label{main} 
% \end{figure*}

\subsection{Federated Prior Knowledge Distillation}
\label{fpkd-section}
%Existing FD methods \cite{he2020group,wu2022exploring} simply fit local predictions to global knowledge during local distillation process, so as to mine the rich structured inter-class similarity information and prevent the local models from overfitting \cite{hinton2015distilling}.
Existing FD methods \cite{he2020group,wu2022exploring} without public datasets simply let local models fit downloaded global knowledge during local distillation, during which all local models learn a uniform global representation, which is commonly generalized and relatively class-balanced.
However, in FMTL, the training tasks of local models are highly correlated with local data distributions, and more biased local representation is preferred.
Thus, we optimize client-side local models utilizing local data distributions and concentrate on classes with high frequencies to adapt to skewed local data.
%Therefore, we utilize local data distributions to guide local models' optimization and make client-side local models focus on classes with high frequencies \textcolor{black}{to adapt to skewed local data.}
%In this paper, we utilize local data distributions to guide local models' optimization: focusing more on classes with high local frequencies. 
%to enhance the preference of global knowledge for major local categories and to improve the fitting of clients to local data.
Specifically, for the $i$-th sample of client $k$ denoted as $\hat X^k_i$, the $r$-th dimension of its global knowledge is denoted as $global_r:=(z_{\hat X_i^k}^S)_r$, and the $r$-th dimension of its local knowledge is denoted as $local_r:=(z_{\hat X_i^k}^k)_r$.
In addition, $w^k_i$ is defined to weight the $i$-th component of KL-divergence between the local knowledge of client $k$ and the global knowledge. %enabling local models to prefer high-frequency local categories rather than simply fit the global knowledge.
%rather than directly fitting the global knowledge.
Accordingly, the optimization objective of client $k$ is defined as follows:
\begin{equation}
	J_{FPKD}^k({W^k};{d^k}) = \mathop {{\rm{ }}E}\limits_{(\hat X_i^k,\hat y_i^k)\sim{{\hat {\cal D}}^k}} [  \sum\limits_{r = 1}^C {w_r^k \cdot {global_r}}\cdot \log \frac{{{global_r}}}{{{local_r}}}],
	\label{lga-client-fpkd1}
\end{equation}
where $w^k_r$ is positively correlated to local class frequencies and is controlled by a hyperparameter $T$, that is:
\begin{equation}
	w_r^k = \frac{{{\exp({\frac{{f_r^k}}{T}})}}}{{\sum\limits_{j = 1}^C {{\exp({\frac{{f_j^k}}{T}})}} }},
	\label{lga-client-pfkd2}
\end{equation}
where $f_i^k$ denotes the sample frequency of category $i$ in $\hat {\cal D}^k_i$.

\subsection{Local Knowledge Adjustment}
%For FL, differentiated local learning tasks will result in gradually forgetting representations of global models
An essential issue of noteworthy divergence among local models needs to be solved during global distillation in FMTL, deriving from data heterogeneity and personalized local distillation (e.g., FPKD discussed in section \ref{fpkd-section}).
%FMTL faces an essential issue of the significant global-local model divergence resulting from data heterogeneity as well as local adaptation techniques used to enhance multi-task capabilities of clients (e.g., FPKD discussed in section \ref{fpkd-section}).
%On the one hand, local models tend to learn discriminative representations depending on local data rather than the overall data.
%On the other hand, our proposed FPKD further increases these biases since it enhances the multi-task capabilities of local models.
Recent works have demonstrated that local divergence is detrimental to the overall FL training, as client-side local models tend to gradually forget representations of global models and drift towards their local objectives \cite{lee2021preservation,he2022class}.
This phenomenon inevitably poses inconsistent updates and unstable convergence when aggregating highly-differentiated local models, i.e. client drift \cite{lee2021preservation,he2022class,karimireddy2020SCAFFOLD,yao2021local}.
To this end, we expect to tackle the above-mentioned problem by assigning importance to local knowledge.
%To this end, we expect to tackle the above problem by alleviating the negative effects derived from local model discrepancies on global distillation, aiming to guarantee that local knowledge matches the global representation.
Specifically, we consider two levels:% of local knowledge as follows: 
\begin{itemize}
    \item \textbf{Client level.} The global model optimization pays more attention to local knowledge from clients whose local data distributions are similar to the overall data distribution.
    \textcolor{black}{As a result, the server's collaboration with clients whose private data distribution is similar to overall data distribution is strengthened, reducing inter-relevant knowledge transfer from clients.}
    \item \textbf{Class level.} The class importance in global distillation is positively correlated with the residuals of global-local class frequencies. \textcolor{black}{This technique balances local information across classes to avoid the global model from learning biased local class representations.}
\end{itemize}
%Techniques based on the above two considerations are respectively named similarity-based and class-balanced LKA, and will be elaborated on in the following subsections.
Based on the above-mentioned two insights, we propose similarity-based and class-balanced LKA respectively. They will be elaborated on in the following subsections.

%In this paper, we tackle the above problem by correcting the distillation loss on the server based on the overall data distribution and the local data distributions of clients who upload their knowledge. 
%Specifically, two techniques from different perspectives are proposed: controlling the importance of knowledge from different classes to balance the different preferences for the same category across clients; or focusing more on distilling knowledge from clients with similar data distributions to improve the generalization of the global model. The proposed techniques will be elaborated in the following subsections.

\subsubsection{Similarity-based Local Knowledge Adjustment}
%Reinforce the collaboration between clients with similar data distributions facilitates personalized accuracy in FD \cite{cho2021personalized,zhang2021parameterized}.
The training performance of FD can be improved through knowledge collaboration among clients with similar data distributions, as pointed out in \cite{cho2021personalized,zhang2021parameterized}. 
Likewise, global distillation can be enhanced with the collaboration of clients whose data distributions are similar to overall data distribution.
Hence, we design distribution-wise weights on local knowledge, aiming to reduce the negative effects of inconsistent knowledge on the global model.
%we enhance the server's collaboration with clients whose data distributions match the overall data distribution by applying distribution-wise adaptive weights on local knowledge, aiming that the global model learns less irrelevant knowledge and improves generalization.
%Inspired by this insight, we achieve effective server-client collaboration by adopting distribution-wise adaptive weights. 
%to help filter out knowledge whose latent local data distribution is similar to the overall data distribution.
Precisely, the similarity difference between global and local knowledge is measured by the cosine similarity of global and local data distribution vectors.
Then, the weights of local knowledge from clients are proportional to the resulting knowledge similarity during global distillation.
%, which enables the server to focus more on the clients whose local data distributions are similar to the overall data distribution.
%we extend the methodologies to server-client collaboration: the server adopts distribution-wise adaptive weights to help capture the similarity of global and local data distributions, and the server will focus more on learning the knowledge of the clients whose distributions of local data is similar to that of overall data. Specifically, the loss of global and local knowledge similarity is weighted by the cosine similarity of global and local data distribution vectors. 
The global distillation objective based on data distribution similarity is defined as follows:
\begin{equation}
	\begin{array}{l}
		\; \; \; \; J_{LKA}^S({W^k};{d^S},{d^k}) \\
		= \mathop E\limits_{k \in \mathcal{K}} \{ \frac{{{(d^S)^\top} \cdot {d^k}}}{{\|{d^S}\|_2 \cdot \|{d^k}\|_2}}\cdot \mathop E\limits_{(\hat X_i^k,\hat y_i^k)\sim{{ \hat  \mathcal{D}}^k}} [L_{sim}(global,local)]\}.
	\end{array}
	\label{lka-sim-new}
\end{equation}

\subsubsection{Class-balanced Local Knowledge Adjustment}
Due to different user behaviors, local data is often class-unbalanced in FL scenarios \cite{shang2022fedic}.
As a result, local model training on each client is strongly correlated with local class distributions and naturally pays more attention to high-frequency categories. 
Not only because high-frequency categories are assigned higher probabilities to reduce the local loss, but also because FPKD enhances local data fitting degrees of local models.
This phenomenon hampers global distillation and slows down model convergence.
%In practical FL scenarios, local data is commonly class-unbalanced due to differences in user behaviors \cite{shang2022fedic}.
%As a result, local model training on each client is strongly correlated with class distributions and naturally pays more attention to high-frequency categories: not only because the model tends to assign a higher probability to high-frequency categories to reduce overall local loss, but also caused by the above-mentioned FPKD who enhances local-fitting degrees of clients.
%This phenomenon inevitably increases the difficulty of global distillation and further slows down model convergence. 
%Without intervention, local models tend to categorize data into the most common classes, not only because the model tends to assign a higher probability to high-frequency categories to reduce local loss, but also due to the adaptation to local data distributions achieved by FPKD. As it appears unstabile of hidden preference information of local knowledge across clients, directly distilling on such knowledge can lead to slow convergence on the global model.
To this end, we propose a soft-label weighting technique based on class frequency residuals, which assigns lower weights to classes whose local class frequencies on clients are higher than global class frequencies during global distillation. 
This technique can narrow global-local knowledge discrepancy by balancing the transferred local knowledge among classes, preventing the global model from learning skewed local class representations.
%so as to balance the transferred local knowledge among classes, narrowing the  between local and global knowledge and preventing the global model from learning locally skewed class representations.
%In this subsection, we correct local knowledge to better fit the overall data distribution: a soft-label weighting method based on frequency differences is adopted to give lower weights to classes whose local data frequencies are higher than the corresponding class frequencies of overall data, so as to alleviate the differences between the structured information of local and global knowledge. 
The global distillation objective based on class importance is defined as follows:
\begin{equation}
	\begin{array}{l}
			\; \; \; \; J_{LKA}^S({W^k};{d^S},{d^k}) \\
			= \mathop {{\rm{ }}E}\limits_{k \in \mathcal{K}} \{ \mathop {{\rm{ }}E}\limits_{(\hat X_i^k,\hat y_i^k)\sim{{\hat {\cal D}}^k}} [  \sum\limits_{r = 1}^C {v_r^k \cdot {local_r}}\cdot \log \frac{{{local_r}}}{{{global_r}}}]\} ,
	\end{array}
	\label{lka-balance-new1}
\end{equation}
where $v^k_r$ is positively related to the residuals between the global and local class frequencies and is controlled by a hyperparameter $U$, that is:
\begin{equation}
	v_r^k = \frac{{\exp (\frac{{f_r^S - f_r^k}}{U})}}{{\sum\limits_{j = 1}^C {\exp (\frac{{f_j^S - f_j^k}}{U})} }},
	\label{lka-balance-new2}
\end{equation}
where $f_i^S$ denotes the sample frequency of category $i$ in $\bigcup\limits_{k \in \mathcal{K}} {{\hat \mathcal{D}}_i^k}$.

\subsection{Formal Description of FedICT}
The proposed FedICT on clients and the server are illustrated in algorithms \ref{alg1} and \ref{alg2} respectively, where $\bm{H}^k:=\bigcup\limits_{i = 1}^{{N^k}} {\hat H_i^k}$, ${\bm{Y}^k}: = \bigcup\limits_{i = 1}^{{N^k}} {\hat y_i^k}$, $\bm{Z}_{{{\hat X}^k}}^k: = \bigcup\limits_{i = 1}^{{N^k}} {z_{\hat X_i^k}^k}$, $\bm{Z}_{{{\hat X}^k}}^S: = \bigcup\limits_{i = 1}^{{N^k}} {z_{\hat X_i^k}^S}$ and other notations are listed in TABLE \ref{notation}. 
To start with, $K$ clients and the server simultaneously execute their corresponding algorithms, where clients start execution by calling FedICT-CLIENT (Algorithm \ref{alg1}, line 1), and the server starts by calling FedICT-SERVER (Algorithm \ref{alg2}, line 1).

\begin{algorithm}[b]
	\begin{spacing}{1.0}
		\SetAlgoNoLine
		\caption{FedICT on Client $k$.
		}
		\begin{algorithmic}[1]		
			\Procedure{FedICT-Client}{${{\hat {\cal D}^k}}$, ${W^k}$, $N^k$} %\Comment{Call entry}
			\State $d^k$= \Call{LocalInit}{${\hat {\cal D}^k}$, $N^k$}
			\State \Repeat{Reaches communication rounds $R$}{
				\State \space $W^k$=\Call{LocalDistill}{${{\hat {\cal D}^k}}$, ${W^k}$, $d^k$}
			}
			\State \Return{Trained $W^k$}
			\EndProcedure
			
			\Procedure{LocalInit}{${\hat {\cal D}^k}$, $N^k$}
			\State Compute $d^k$ according to Eq. (\ref{dist-k})
			\State Upload $d^k$, $N^k$ and $\bm{Y}^k$ to the server
			\State \Return{$d^k$}
			\EndProcedure

			\Procedure{LocalDistill}{${{\hat {\cal D}^k}}$, ${W^k}$, $d^k$}
			\State Receive $\bm{Z}^S_{{\hat X}^S}$ from the server
			\State Optimize $J_{ICT}^k$ according to Eq. (\ref{clinet-lga-gener})
			\State Extract ${\bm{H}}^k$ according to Eq. (\ref{H-k})
			\State Extract $\bm{Z}^k_{{\hat X}^k}$ according to Eq. (\ref{z-k})
			\State Upload $\bm{H}^k$ and $\bm{Z}^k_{{\hat X}^k}$ to the server
			\State \Return Trained $W^k$
			\EndProcedure
		\end{algorithmic}
		\label{lga-client}
		\label{alg1}
	\end{spacing}
\end{algorithm}

\begin{algorithm}[t]
	\begin{spacing}{1.0}
		\SetAlgoNoLine
		\caption{FedICT on the Server. }
		\begin{algorithmic}[1]

			\Procedure{FedICT-Server}{${W^S}$} %\Comment{Call entry}
			\State $d^S$,$\bigcup\limits_{k = 1}^K {{d^k}} $, $\bigcup\limits_{k = 1}^K {{\bm{Y}^k}} $=\Call{GlobalInit()}{}
			\State \Repeat{Reaches communication rounds $R$}
			{
				\State \space $W^S$= \Call{GlobalDistill}{$W^S$,$d^S$,$\bigcup\limits_{k = 1}^K {{d^k}} $,$\bigcup\limits_{k = 1}^K {{\bm{Y}^k}} $}
			}
			\State \Return Trained $W^S$
			\EndProcedure
			\Procedure{GlobalInit()}{}
			\State \space Receive all $d^k$, $N^k$ and $\bm{Y}^k$ from clients
			\State \space Compute ${d^S} = {{\sum\limits_{k = 1}^K {{N^k} \cdot {d^k}} }}\bigg/{{\sum\limits_{k = 1}^K {{N^k}} }}$
			\State \ForAll{Client k}
			{
				\State \space Initialize ${\bm{Z}^S_{{\hat X}^k}}$ with zeros
				\State \space Distribute ${\bm{Z}^S_{{\hat X}^k}}$ to client $k$
			}
			\State \Return $d^k$, $\bigcup\limits_{k = 1}^K {{d^k}} $,$\bigcup\limits_{k = 1}^K {{\bm{Y}^k}} $
			\EndProcedure
			
			\Procedure{GlobalDistill}{$W^S$,$d^S$,$\bigcup\limits_{k = 1}^K {{d^k}} $,$\bigcup\limits_{k = 1}^K {{\bm{Y}^k}} $}
			\State \ForAll{Client k}
			{
				\State \space Receive $\bm{H}^k$ and $\bm{Z}^k_{{\hat X}^k}$ from client $k$
				\State \space Optimize $J_{ICT}^S$ according to Eq. (\ref{server-lga-gener})
				\State \space Generate ${\bm{Z}^S_{{\hat X}^k}}$ according to Eq. (\ref{z-S})
				\State \space Distribute ${\bm{Z}^S_{{\hat X}^k}}$ to client $k$
			}
			\State \Return Trained $W^S$
			\EndProcedure
			
		\end{algorithmic}
		\label{lga-server}
		\label{alg2}
	\end{spacing}
\end{algorithm}

%FedICT performs one global distillation on the server and one local distillation on all clients  at each training round.
All clients first perform local initialization (Algorithm \ref{alg1}, line 2) as follows: clients parallelly compute their local data distribution vectors based on Eq. (\ref{dist-k}) (Algorithm \ref{alg1}, line 7). After that, the local data distribution vectors, local sample numbers and local labels are sent to the server (Algorithm \ref{alg1}, line 8), followed by iteratively conducting local distillation (Algorithm \ref{alg1}, line 4). 
Meanwhile, the server first performs global initialization (Algorithm \ref{alg2}, line 2), which includes receiving the local data information from all clients (Algorithm \ref{alg2}, line 7) and then calculating the global data distribution vector (Algorithm \ref{alg2}, line 8).
After that, the server sets the global knowledge to zeros and distributes the initialized values to all clients (Algorithm \ref{alg2}, lines 9-11). Subsequently, the server iteratively performs global distillation until training stops (Algorithm \ref{alg2}, line 4). 
%Each time when both global distillation on the server and local distillation on all clients are complete for one iteration, FedICT performs a communication round.

At the beginning of each training round, all clients parallelly receive the global knowledge generated by the server in the previous round (Algorithm \ref{alg1}, line 11). The local model parameters are then optimized according to Eq. (\ref{clinet-lga-gener}), during which the prior knowledge about clients' local data distributions is injected to guide local models to accommodate their local data (Algorithm \ref{alg1}, line 12). 
Subsequently, local knowledge is extracted and uploaded to the server (Algorithm \ref{alg1}, lines 13-15). The server then accepts the local knowledge uploaded by each client (Algorithm \ref{alg2}, line 15) and optimizes the global model parameters according to Eq. (\ref{server-lga-gener}) (Algorithm \ref{alg2}, line 16). 
%,\ref{lka-sim1},\ref{lka-sim2},\ref{lka-balance}
%It is worth noting that this step requires customizing the optimization constraints for global distillation in advance, where adopting either class-balanced or similarity-based local knowledge adjustment is acceptable. 
Noting that this operation benefits global distillation via similarity-based LKA according to Eq. (\ref{lka-sim-new}) or class-balanced LKA according to Eq. (\ref{lka-balance-new1}).
Further, the server extracts the global knowledge based on the updated global model parameters and distributes them to corresponding clients (Algorithm \ref{alg2}, lines 17-19). 
The whole training process is completed until model convergence.

\begin{figure*}[t!]
	\centering
	\includegraphics[width=0.95 \textwidth]{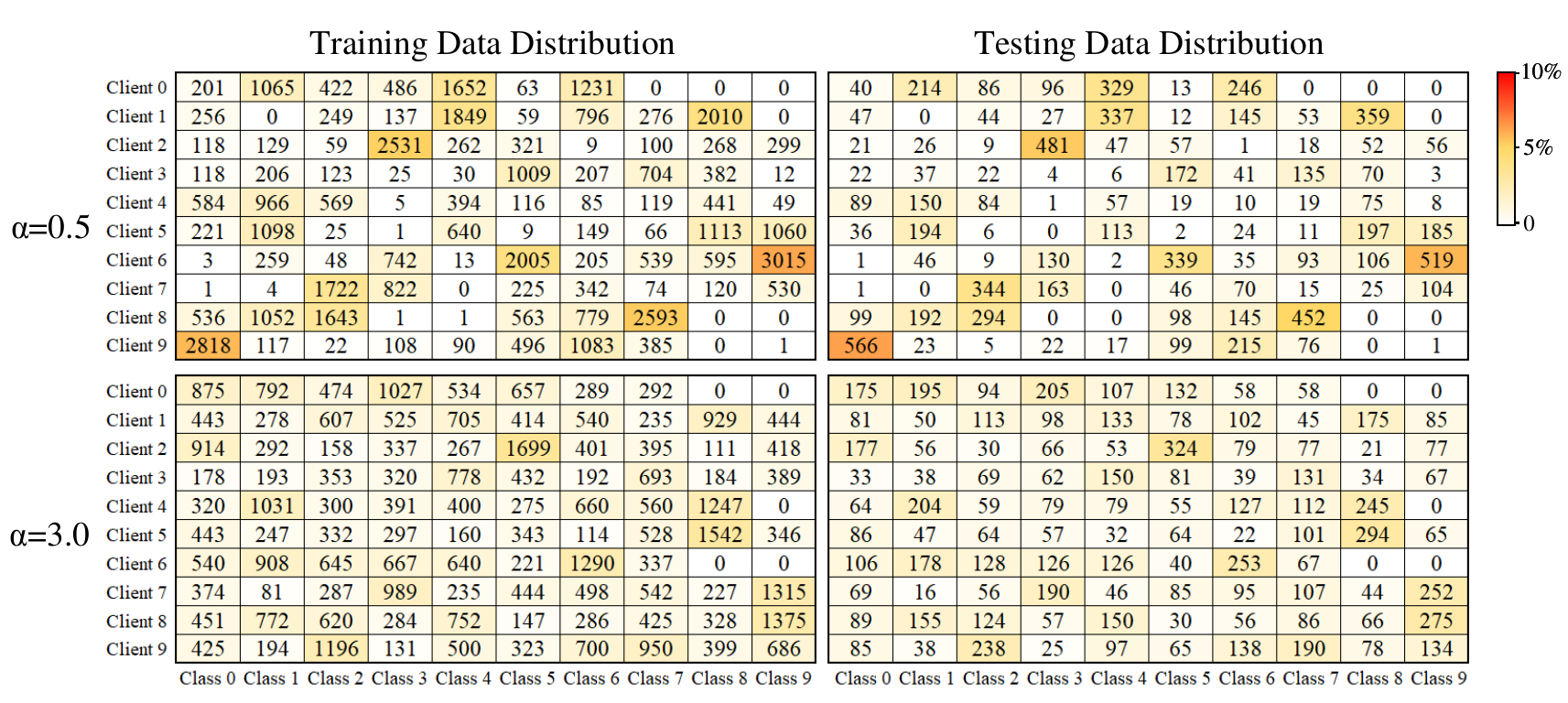}
	\caption{Data distributions with different $\alpha$ on CIFAR-10. Each heat map represents the training/testing data distributions for all clients. Each row of heat maps represents the class distributions of a single client, where the column label gives the category. Each cell represents the sample number of corresponding classes for a given client's training/testing dataset, and the shade of the color indicates the proportion to the total. } %caption是图片的标题
	\label{data-dist} 
\end{figure*}

\section{Experiments}
\label{experiments}
\subsection{Experimental Setup}
\subsubsection{Datasets and Preprocessing}
%We compare our methods with state-of-the-arts both in simulation experiments on CIFAR-10 \cite{krizhevsky2009learning},  CINIC-10 \cite{darlow2018cinic} for image classification, and in mobile application experiments on TMD \cite{carpineti2018custom} for transportation mode detection. 
\textbf{Datasets.} We conduct experiments on image datasets CIFAR-10 \cite{krizhevsky2009learning}, CINIC-10 \cite{darlow2018cinic} for classification, and one mobile sensor data mining dataset TMD \cite{carpineti2018custom} for transportation mode detection. CIFAR-10 and CINIC-10 are 10-class image classification datasets with common objects. TMD is a 5-class transportation mode detection dataset that categorizes heterogeneous users' transportation modes by mining embedded sensor data from smartphones. All datasets are pre-split into training and testing datasets.

\noindent
\textbf{Data Partition.} For all of our experiments, data partitioning strategy in \cite{he2020fedml} is adopted, where the hyper-parameter $\alpha$ ($\alpha>0$) controls the degree of data heterogeneity, with a smaller $\alpha$ indicating a stronger degree of heterogeneity. 
In the FMTL setup, the testing dataset of each client satisfies a similar distribution with its training dataset. 
Fig. \ref{data-dist} shows the data distributions of training/testing datasets on CIFAR-10 when 10 clients participate in FMTL. 
As displayed, the heat map with the smaller $\alpha$ exhibits more uneven color distributions, i.e., more unbalanced data partition. Moreover, the color distributions of training and testing datasets for each client are almost identical, i.e., isomorphic training/testing data distribution for individual clients.
For experiments on image classification, we conduct two groups of experiments under conditions of homogeneous and heterogeneous models, each with 10 and 5 clients, respectively. Each experiment group validates on three different degrees of data heterogeneity, $\alpha \in \{0.5, 1.0, 3.0\}$. 
For experiments on transportation mode detection, we respectively set the numbers of participated devices to 120 and 150 under two data heterogeneity settings, $\alpha\in\{1.0,3.0\}$.
%hence corresponding to the more non-independent and identically distributed (NonIID) partitioned dataset. 
%therefore, they are independent and identically distributed (IID). 

\noindent
\textbf{Data Augmentation and Normalization.} 
%For simulation experiments, we conduct two groups of experiments in the client-side model homogeneous and heterogeneous conditions, respectively. 
%Specifically, each dataset is partitioned into 10 and 5 parts to respectively simulate 10 and 5 clients in experiments on homogeneous and heterogeneous models.
%For experiments on homogeneous local models, each dataset is partitioned into 10 parts to simulate 10 clients, while with the same model architecture. For experiments on heterogeneous models, each dataset is partitioned into 5 parts to simulate 5 clients with five given model architectures. 
For experiments on image classification, we conduct random crop, random horizontal flip and mean-variance standardization before feeding images into models. For experiments on transportation mode detection, we normalize the sensor data to have a mean of 0 and a variance of 1.
%For experiments on mobile applications, we respectively consider 120 and 150 devices in FL training and validate our methods with two data NonIID settings, $\alpha\in\{1.0,3.0\}$. All input sensor information is normalized to have a mean of 0 and a variance of 1.

\begin{table}[]
\renewcommand\arraystretch{1.1}
	\caption{Main configuration of models. $H$ and $W$ are the height and width of input images, respectively.}
	\centering
\begin{tabular}{c|c|c|c}
\hline
\textbf{Notation} & \textbf{Type}                                                                                & \textbf{Feat. Shape}            & \textbf{Params} \\ \hline
$A_1^C$         & \multirow{5}{*}{\begin{tabular}[c]{@{}c@{}}Convolutional\\ Neural\\ Network\end{tabular}} & \multirow{5}{*}{$H \times W \times 16$} & 0.7K   \\
$A_2^C$         &                                                                                           &                         & 5.2K   \\
$A_3^C$         &                                                                                           &                         & 10.5K  \\
$A_4^C/A_5^C$      &                                                                                           &                         & 9.8K   \\
$A_1^S$        &                                                                                           &                         & 588.2K \\ \hline
$A_6^C$         & \multirow{4}{*}{\begin{tabular}[c]{@{}c@{}}Fully Connected\\ Neural Network\end{tabular}} & \multirow{4}{*}{13}     & 1109   \\
$A_7^C$         &                                                                                           &                         & 1335   \\
$A_8^C$         &                                                                                           &                         & 1877   \\
$A_2^S$        &                                                                                           &                         & 2053   \\ \hline
\end{tabular}
\label{model-conf}
\end{table}

\subsubsection{Models}
\label{exp-models}
In our experiments, a total of eight local model architectures $\{A^C_1,......, A^C_8\}$ are adopted, wherein $\{A^C_1,......,A^C_5\}$ are convolutional neural networks for image classification, and $\{A^C_6,A^C_7,A^C_8\}$ are fully connected neural networks for transportation mode detection. 
In particular, global model architectures $A^S_0$ and $A^S_1$ are adopted for image classification and transportation mode detection, respectively. 
Details of model configurations are provided in TABLE \ref{model-conf}.
For image classification experiments with homogeneous models, all clients adopt the same model architecture $A^C_1$. For image classification experiments with heterogeneous models, each of the five clients adopts a different model architecture $\{A^C_1,......,A^C_5\}$. 
In transportation mode detection experiments, we randomly choose $A^C_8$ architecture with a 10\% probability, $A^C_7$ architecture with a 30\% probability, and $A^C_6$ architecture for the rest when adopting FD methods. 
For clients adopting non-FD methods, we conduct three groups of experiments with different model architectures, in which $A^C_6$, $A^C_7$ and $A^C_8$ are respectively adopted for all clients in each group.

\begin{table*}[t]
	\centering
	\renewcommand\arraystretch{1.1}
	\setlength{\tabcolsep}{10pt}
	\caption{Average UA (\%) \cite{mills2021multi} on homogeneous local models. \textbf{Bold} values represent the best performance, and {\ul underlined} values represent the second-best performance. The same as below. }
	\begin{tabular}{l|c|ccc|ccc}
		\hline
		\multicolumn{1}{c|}{\multirow{2}{*}{\textbf{Method}}} & \multirow{2}{*}{\textbf{Model}} & \multicolumn{3}{c|}{\textbf{CIFAR-10}}            & \multicolumn{3}{c}{\textbf{CINIC-10}}            \\
		\multicolumn{1}{c|}{}                                 &                                 & \textbf{$\bm{\alpha}$=3.0} & \textbf{$\bm{\alpha}$=1.0} & \textbf{$\bm{\alpha}$=0.5} & \textbf{$\bm{\alpha}$=3.0} & \textbf{$\bm{\alpha}$=1.0} & \textbf{$\bm{\alpha}$=0.5} \\ \hline
		FedAvg                                                & \multirow{9}{*}{$A^C_1$}             & 45.73          & 39.97          & 38.28          & 45.76          & 42.06          & 39.30          \\
		FedAdam                                               &                                 & 49.09          & 53.03          & 40.13          & 55.71          & 54.03          & 49.72          \\
		pFedMe                                                &                                 & 37.53          & 34.78          & 32.73          & 41.03          & 38.33          & 34.59          \\
		MTFL                                                  &                                 & 42.59          & 38.99          & 36.96          & 42.60          & 39.32          & 35.67          \\
              \textcolor{black}{DemLearn} &
              & 35.35          & 37.20          & 46.61          & 32.87    & 35.76           & 45.44 \\
		FedGKT                                                &                                 & 59.34          & 63.83          & 71.26          & 46.96          & 48.58          & 57.56          \\
		FedDKC                                                &                                 & 60.30          & 62.70          & 71.53          & 50.92          & 51.35          & 61.09          \\
		\textbf{FedICT (sim)}                                  &                                 & {\ul 60.96} & \textbf{65.42}    & \textbf{73.54} & \textbf{56.49}    & {\ul 57.05}    & {\ul 65.46}    \\
		\textbf{FedICT (balance)}                              &                                 & \textbf{61.28}    & {\ul 65.15} & {\ul 73.37}    & {\ul 56.34} & \textbf{57.12} & \textbf{65.72} \\ \hline
	\end{tabular}
	\label{exp-homo}
\end{table*}

\subsubsection{Benchmarks}
We compare FedICT combined with FPKD and LKA with state-of-the-art methods as follows:
\begin{itemize}
	\item
	Classical FL method, FedAvg \cite{mcmahan2017communication} and FedAdam \cite{reddi2021adaptive}.
	%\item
	%Fast convergence FL method, FedAdam \cite{reddi2021adaptive}.
	\item
	Personalized FL method, pFedMe \cite{t2020personalized}.
	\item
	FMTL method, MTFL \cite{mills2021multi}.
	\item
        \textcolor{black}{Multi-task distributed learning method, DemLearn \cite{nguyen2022self}}
        \item
	FD methods, FedGKT \cite{he2020group} and FedDKC \cite{wu2022exploring}.
\end{itemize}
Of all the above methods, FD methods support heterogeneous local models, while non-FD methods only support homogeneous local models. Hence, in image classification experiments, we compare FedICT with all the above state-of-the-art methods on homogeneous models, while only compare FedICT with FD methods on heterogeneous models.
In experiments on transportation mode detection, we simultaneously compare our proposed methods with all the above benchmarks, where FD-based methods adopt heterogeneous models with random model architectures, and non-FD methods respectively adopt three different model architectures, as discussed in section \ref{exp-models}.
Moreover, we adopt average User model Accuracy (UA) as the evaluation metric referred to \cite{mills2021multi}, where UA denotes the training accuracy of client-side local models through validating on local testing datasets.

\subsubsection{Hyper-parameter Settings}
We adopt stochastic gradient descent to optimize all models. For experiments on image classification, we set the learning rate to $1 \times 10^{-2}$, the $l_2$ weight decay value to $5\times 10^{-4}$, and the batch size to 256. For experiments on transportation mode detection, the learning rate, weight decay value, and batch size are set as $3 \times 10^{-4}$, $5\times 10^{-4}$ and 2, respectively.
For all the compared methods, each client optimizes its local model for an epoch before conducting parameter aggregation or global distillation.
%we require clients to optimize locally for the overall epoch before conducting parameter aggregation or global distillation. 
Some methods require individualized hyper-parameters, which are set as follows:
\begin{itemize}
	\item
	We set $\beta_1=0.9$, $\beta_2=0.99$ and $\tau=0.001$ in FedAdam referencing to \cite{reddi2021adaptive}.
	\item
	We set $\eta=0.005$, $\lambda=15$, $\beta=1$ in pFedMe, referencing to \cite{t2020personalized}.
	\item
 We adopt implementation based on FedAvg in MTFL, with other hyper-parameters kept as default in \cite{mtflurl}.
	%To compare with MTFL, we adopt implementation based on FedAvg, with other hyper-parameters kept as default in \cite{mtflurl}.
        \item
        \textcolor{black}{We adopted the default hyper-parameter settings \cite{demurl} in DemLearn.}
	\item
	We adopt the empirically more effective scheme, KKR-FedDKC, with $\beta=1.5$ and $T=0.12$ referencing to \cite{wu2022exploring}.
%We set $\beta=1.5$ and $T=0.12$
 %The hyper-parameter settings of FedDKC is referred to \cite{wu2022exploring} with $\beta=1.5$ and $T=0.12$.
	\item
	We set $\beta=\lambda=\mu=1.5$, $T=3.0$ and $U=7.0$ in our proposed FedICT.
\end{itemize}
%The notations in each of the above items are the hyper-parameters in the paper proposing corresponding methods.
%In all of our experiments, the individualized hyper-parameters keep consistent to avoid excessive adjustment.

\begin{table*}[h]
	\renewcommand\arraystretch{1.1}
	\setlength{\tabcolsep}{10pt}
	\caption{UA (\%) on heterogeneous local models.}
	\centering
	\begin{tabular}{l|c|ccc|ccc}
		\hline
		\multicolumn{1}{c|}{\multirow{2}{*}{\textbf{Method}}} & \multirow{2}{*}{\textbf{Model}} & \multicolumn{3}{c|}{\textbf{CIFAR-10}}           & \multicolumn{3}{c}{\textbf{CINIC-10}}            \\
		\multicolumn{1}{c|}{}                                 &                                 & \textbf{$\bm{\alpha}$=3.0} & \textbf{$\bm{\alpha}$=1.0} & \textbf{$\bm{\alpha}$=0.5} & \textbf{$\bm{\alpha}$=3.0} & \textbf{$\bm{\alpha}$=1.0} & \textbf{$\bm{\alpha}$=0.5} \\ \hline
		\multirow{6}{*}{FedGKT}                               & $A^C_1$                              & 35.55          & 44.62          & 49.90          & 39.95          & 48.82          & 52.21          \\
		& $A^C_2$                              & 52.97          & 59.09          & 56.67          & 43.14          & 49.84          & 56.97          \\
		& $A^C_3$                              & 61.04          & 67.15          & 70.16          & 62.75          & 59.40          & 65.84          \\
		& $A^C_4$                              & 50.30          & 54.20          & 68.89          & 45.15          & 43.24          & 62.21          \\
		& $A^C_5$                              & 57.98          & 58.79          & 55.49          & 55.05          & 53.21          & 63.35          \\
		& Clients Avg.                            & 51.57          & 56.77          & 60.22          & 49.21          & 50.90          & 60.12          \\ \hline
		\multirow{6}{*}{FedDKC}                               & $A^C_1$                              & 39.63          & 46.83          & 51.90          & 42.47          & 52.06          & 52.07          \\
		& $A^C_2$                              & 56.48          & 66.43          & 61.61          & 46.66          & 56.43          & 59.41          \\
		& $A^C_3$                              & \textbf{66.68} & 70.33          & 70.20          & 65.35          & 67.07          & 66.51          \\
		& $A^C_4$                              & 56.37          & 56.86          & 71.23          & 52.72          & 50.13          & 62.44          \\
		& $A^C_5$                              & 64.86          & 62.41          & 61.77          & 62.67          & {\ul 59.73}    & 64.09          \\
		& Clients Avg.                            & 56.08          & 60.57          & 63.34          & 53.97          & 57.08          & 60.90          \\ \hline
		\multirow{6}{*}{\textbf{FedICT (sim)}}                & $A^C_1$                              & {\ul 42.40}    & {\ul 49.77}    & {\ul 54.44}    & {\ul 42.62}    & \textbf{54.03} & \textbf{55.42} \\
		& $A^C_2$                              & \textbf{59.85} & \textbf{68.62} & {\ul 70.01}    & \textbf{48.18} & {\ul 57.42}    & {\ul 67.74}    \\
		& $A^C_3$                              & 66.56          & \textbf{72.63} & {\ul 74.37}    & {\ul 65.92}    & {\ul 67.65}    & {\ul 67.32}    \\
		& $A^C_4$                              & {\ul 59.18}    & {\ul 60.74}    & \textbf{73.57} & \textbf{56.13} & {\ul 52.81}    & \textbf{69.58} \\
		& $A^C_5$                              & {\ul 69.99}    & \textbf{63.54} & {\ul 66.49}    & \textbf{66.27} & \textbf{61.51} & \textbf{66.79} \\
		& Clients Avg.                            & {\ul 59.60}    & {\ul 63.06}    & \textbf{67.78} & {\ul 55.82}    & {\ul 58.68}    & {\ul 65.37}    \\ \hline
		\multirow{6}{*}{\textbf{FedICT (balance)}}            & $A^C_1$                              & \textbf{42.98} & \textbf{50.04} & \textbf{55.06} & \textbf{42.76} & {\ul 53.00}    & {\ul 55.15}    \\
		& $A^C_2$                              & {\ul 57.51}    & {\ul 68.33}    & \textbf{70.20} & {\ul 48.10}    & \textbf{60.15} & \textbf{69.13} \\
		& $A^C_3$                              & {\ul 66.63}    & {\ul 72.46}    & \textbf{74.66} & \textbf{66.97} & \textbf{68.61} & \textbf{67.96} \\
		& $A^C_4$                              & \textbf{61.19} & \textbf{63.02} & {\ul 71.27}    & {\ul 55.70}    & \textbf{53.76} & {\ul 68.56}    \\
		& $A^C_5$                              & \textbf{71.59} & {\ul 62.97}    & \textbf{66.83} & {\ul 65.80}    & 59.70          & {\ul 66.74}    \\
		& Clients Avg.                            & \textbf{59.98} & \textbf{63.36} & {\ul 67.60}    & \textbf{55.87} & \textbf{59.04} & \textbf{65.51} \\ \hline
	\end{tabular}
	\label{model-hetero}
\end{table*}

\begin{table*}[]
	\renewcommand\arraystretch{1.1}
	\setlength{\tabcolsep}{10pt}
	\centering
	\caption{Communication rounds of different FD methods when reaching a given average UA.}
	\begin{tabular}{c|l|cccccc}
		\hline
		\multirow{14}{*}{\textbf{Model Homo.}}   & \multicolumn{1}{c|}{\multirow{3}{*}{\textbf{Method}}} & \multicolumn{6}{c}{\textbf{CIFAR-10}}                                                                        \\
		& \multicolumn{1}{c|}{}                                 & \multicolumn{2}{c}{\textbf{$\bm{\alpha}$=3.0}} & \multicolumn{2}{c}{\textbf{$\bm{\alpha}$=1.0}} & \multicolumn{2}{c}{\textbf{$\bm{\alpha}$=0.5}} \\
		& \multicolumn{1}{c|}{}                                 & \textbf{50\%}    & \textbf{60\%}   & \textbf{50\%}    & \textbf{60\%}   & \textbf{60\%}    & \textbf{70\%}   \\ \cline{2-8} 
		& FedGKT                                                & 101              & 432             & 48               & 161             & 28               & 203             \\
		& FedDKC                                                & 72               & 366             & 37               & 136             & 22               & 189             \\
		& \textbf{FedICT (sim)}                                 & \textbf{42}      & {\ul 212}       & \textbf{23}      & \textbf{92}     & \textbf{18}      & \textbf{95}     \\
		& \textbf{FedICT (balance)}                             & \textbf{42}      & \textbf{208}    & \textbf{23}      & \textbf{92}     & {\ul 19}         & \textbf{95}     \\ \cline{2-8} 
		& \multicolumn{1}{c|}{\multirow{3}{*}{\textbf{Method}}} & \multicolumn{6}{c}{\textbf{CINIC-10}}                                                                        \\
		& \multicolumn{1}{c|}{}                                 & \multicolumn{2}{c}{\textbf{$\bm{\alpha}$=3.0}} & \multicolumn{2}{c}{\textbf{$\bm{\alpha}$=1.0}} & \multicolumn{2}{c}{\textbf{$\bm{\alpha}$=0.5}} \\
		& \multicolumn{1}{c|}{}                                 & \textbf{40\%}    & \textbf{50\%}   & \textbf{40\%}    & \textbf{50\%}   & \textbf{50\%}    & \textbf{60\%}   \\ \cline{2-8} 
		& FedGKT                                                & 15               & -               & 4                & -               & 3                & -               \\
		& FedDKC                                                & 13               & 76              & 3                & 41              & \textbf{2}       & 54              \\
		& \textbf{FedICT (sim)}                                 & \textbf{6}       & \textbf{40}     & \textbf{1}       & {\ul 24}        & \textbf{2}       & \textbf{26}     \\
		& \textbf{FedICT (balance)}                             & \textbf{6}       & \textbf{40}     & \textbf{1}       & \textbf{19}     & \textbf{2}       & \textbf{26}     \\ \hline
		\multirow{14}{*}{\textbf{Model Hetero.}} & \multicolumn{1}{c|}{\multirow{3}{*}{\textbf{Method}}} & \multicolumn{6}{c}{\textbf{CIFAR-10}}                                                                        \\
		& \multicolumn{1}{c|}{}                                 & \multicolumn{2}{c}{\textbf{$\bm{\alpha}$=3.0}} & \multicolumn{2}{c}{\textbf{$\bm{\alpha}$=1.0}} & \multicolumn{2}{c}{\textbf{$\bm{\alpha}$=0.5}} \\
		& \multicolumn{1}{c|}{}                                 & \textbf{50\%}    & \textbf{55\%}   & \textbf{50\%}    & \textbf{55\%}   & \textbf{55\%}    & \textbf{60\%}   \\ \cline{2-8} 
		& FedGKT                                                & 84               & -               & 42               & 94              & 28               & 96              \\
		& FedDKC                                                & 71               & 112             & 30               & 57              & 22               & 70              \\
		& \textbf{FedICT (sim)}                                 & \textbf{42}      & \textbf{80}     & \textbf{18}      & {\ul 43}        & \textbf{13}      & {\ul 43}        \\
		& \textbf{FedICT (balance)}                             & {\ul 45}         & \textbf{80}     & \textbf{18}      & \textbf{42}     & \textbf{13}      & \textbf{41}     \\ \cline{2-8} 
		& \multicolumn{1}{c|}{\multirow{3}{*}{\textbf{Method}}} & \multicolumn{6}{c}{\textbf{CINIC-10}}                                                                        \\
		& \multicolumn{1}{c|}{}                                 & \multicolumn{2}{c}{\textbf{$\bm{\alpha}$=3.0}} & \multicolumn{2}{c}{\textbf{$\bm{\alpha}$=1.0}} & \multicolumn{2}{c}{\textbf{$\bm{\alpha}$=0.5}} \\
		& \multicolumn{1}{c|}{}                                 & \textbf{40\%}    & \textbf{50\%}   & \textbf{50\%}    & \textbf{55\%}   & \textbf{55\%}    & \textbf{60\%}   \\ \cline{2-8} 
		& FedGKT                                                & 8                & 59              & 57               & -               & 37               & -               \\
		& FedDKC                                                & 8                & 61              & 35               & 84              & 15               & 54              \\
		& \textbf{FedICT (sim)}                                 & \textbf{6}       & \textbf{30}     & {\ul 30}         & {\ul 47}        & \textbf{11}      & {\ul 38}        \\
		& \textbf{FedICT (balance)}                             & \textbf{6}       & {\ul 33}        & \textbf{27}      & \textbf{46}     & \textbf{11}      & \textbf{36}     \\ \hline
	\end{tabular}
	\label{communi-FD}
\end{table*}

\begin{figure*}[h]
	\centering
	\includegraphics[width=1.0 \textwidth]{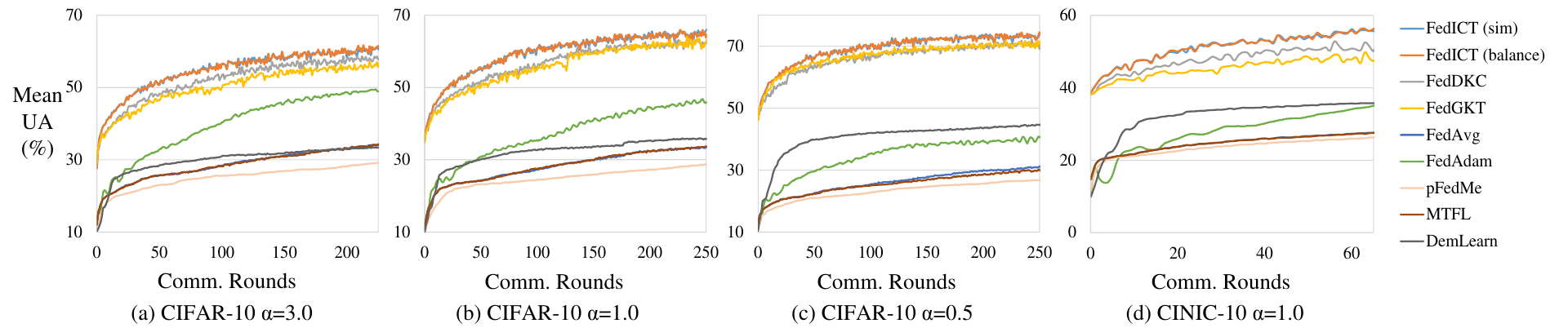}
	\caption{Learning curves of local models measured by average UA on different degrees of data heterogeneity and datasets.}
	\label{conv-datasets} 
\end{figure*}

\subsection{Results on Image Classification}
\subsubsection{Performance on Homogeneous Models}
TABLE \ref{exp-homo} compares our proposed FedICT with existing state-of-the-art methods on two image classification datasets, where all clients adopt the same model architecture $A^C_1$. For the last two lines in the table, we adopt similarity-based LKA in FedICT (sim) and class-balanced LKA in FedICT (balance) , the same as in the following sections. As shown in TABLE \ref{exp-homo}, FedICTs both outperform all other baselines on both CIFAR-10 and CINIC-10 in all data heterogeneity settings. 
Specifically, FedICT (sim) increases the average UA by up to 1.41\% and 2.72\% on CIFAR-10 and CINIC-10 compared with the best performances on six benchmarks respectively, and the improvements are with 1.38\% and 2.78\% for FedICT (balance). 
Hence, we can conclude that our proposed methods are effective in challenging federated multi-task classification with clients' local data exhibiting heterogeneity among each other.

\subsubsection{Performance on Heterogeneous Models}
TABLE \ref{model-hetero} compares the performance of FedICTs with FedGKT and FedDKC, including results on two datasets with three degrees of data heterogeneity and five independently designed models.
We can see that both FedICT (sim) and FedICT (balance) outperform the compared benchmarks in all image classification datasets, all data heterogeneity settings, and all adopted model architectures in terms of the average UA, with more than 3.06\% improvement in average on FedICT (sim), and more than 3.23\% improvement in average on FedICT (balance).
Notably, in the total of 30 client settings, both FedICT (sim) and FedICT (balance) outperform the best performances in FedGKT and FedDKC on 29 clients, i.e., UA's improvement covering 96.67\% of clients.
This result demonstrates that our proposed methods not only improve the average UA of clients, but also are robust to model architectures, which are satisfactory for clients with different data distributions and model architectures.
This property motivates diversified devices with heterogeneous data to participate in FMTL training, and significantly promotes the availability in real MEC scenarios. 
%who motivate to participate in FL training in MEC.

\begin{figure}[t]
	\centering
	\includegraphics[width=0.5 \textwidth]{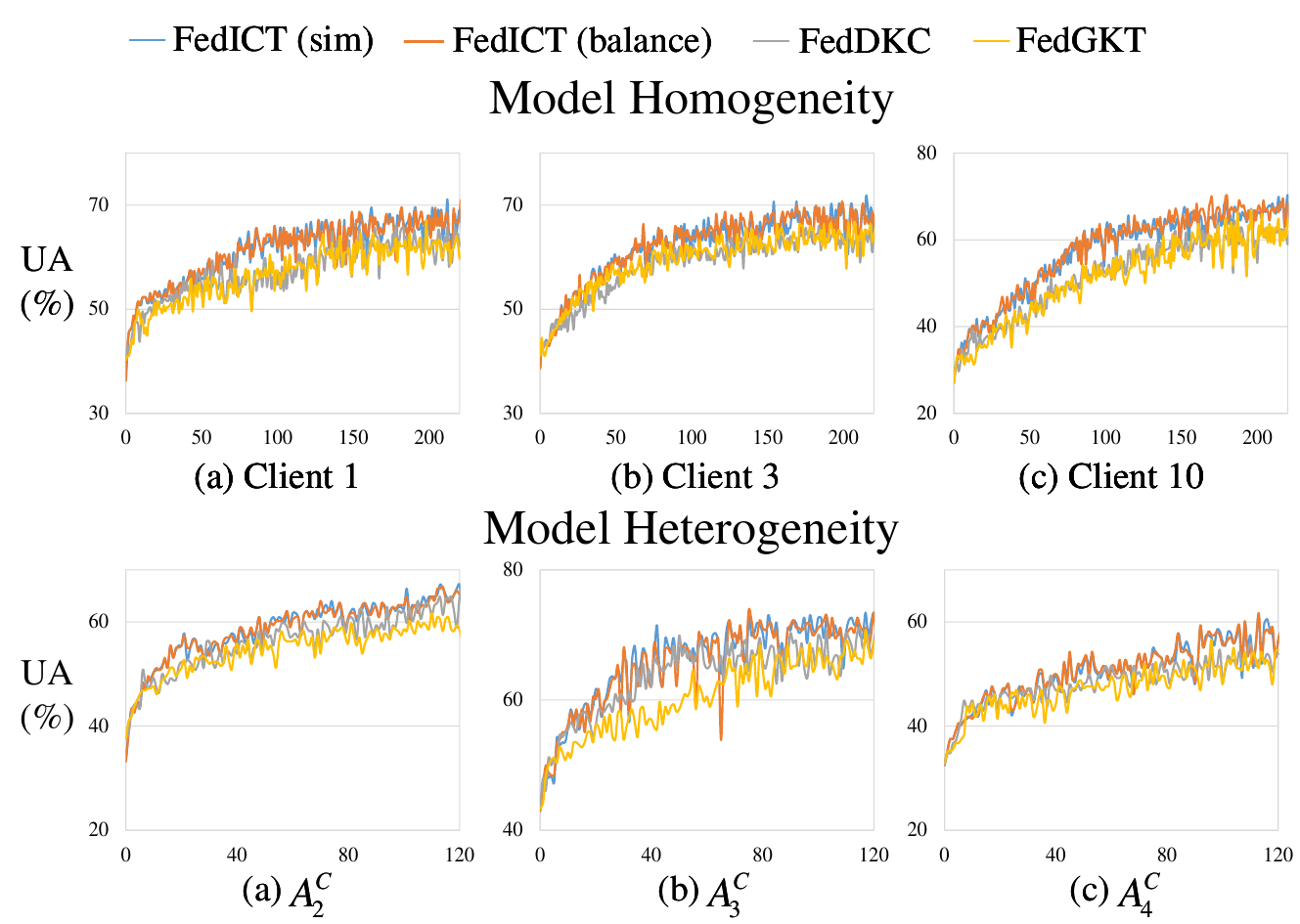}
	\caption{Learning curves on selected local models, where the horizontal coordinates indicate the number of communication rounds. Results are derived from CIFAR-10, taking $\alpha$=1.0.} 
	\label{conv-combine} 
\end{figure}

\subsubsection{Convergence Analysis}
\label{convergence-analysis}
We first suggest that FD methods generally converge much faster than non-FD methods, as displayed in Fig. \ref{conv-datasets}.
Since knowledge and features exchanged in each communication round contain information about multiple rounds of model optimization, FD methods always converge to a higher average UA than non-FD methods under the same number of communication rounds regardless of datasets, model architecture setups, and degrees of data heterogeneity. 
Therefore, we only compare the convergence speed of our proposed FedICTs with existing FD methods by comparing the number of communication rounds required to reach a given average UA. 
As displayed in TABLE \ref{communi-FD}, the required number of communication rounds to converge to all given average UAs for FedICTs are smaller than that of existing FD methods in all settings. 
Specifically, the number of communication rounds required by FedICTs is no more than 75\% of FedGKT to achieve all given average UAs.
Thus, we can draw that FedICTs achieve convergence acceleration, and their training performance suits various data distributions and model architectures.
This is because LKA mitigates client drift derived by local knowledge divergence during global distillation, so the server can capture a more generalizable representation and facilitate local distillation with the assistance of FPKD in turn.

We further confirm the effectiveness of FedICTs in improving the convergence of individual clients.
Fig. \ref{conv-combine} displays the learning curves of selected models under both homogeneous and heterogeneous local model settings. We can figure out that FedICTs consistently exhibit faster convergence compared to FedGKT and FedDKC and can converge to higher UA in all selected clients. This confirms that our proposed methods can improve the convergence performance of heterogeneous individual clients, which supports the fairness of FedICTs for clients under various conditions.

\subsection{Results on Transportation Mode Detection}
TABLE \ref{mobile-table} shows the comparison of FedICTs with all considered state-of-the-art methods on TMD dataset under different model architecture settings. 
We can see that our proposed methods achieve the highest communication efficiency than all benchmarks on both 120 and 150 clients settings, %when $\{120, 150\}$ clients are simulated,
regardless of the degrees of data heterogeneity and model architectures. 
Specifically, benefiting from exchanging only compact features and knowledge between the server and clients, FedICTs require less than 1.2\% and 0.6\% of communication overheads to achieve 37\% average UA in settings of 120 and 150 clients compared with FedAvg. 
%Regarding average UA, FedICTs outperform all benchmark algorithms when setting both 120 and 150 clients. 
This demonstrates that our proposed methods simultaneously achieve efficient communication, allow heterogeneous local models, and enable performance on task-diverse clients superior to state-of-the-art methods, which are not only practical for MEC but also can remarkably improve client-side training accuracy in multi-task settings.

\begin{table*}[]
	\centering
	\renewcommand\arraystretch{1.1}
	\caption{Average UA (\%) and communication overheads on TMD dataset, taking $\alpha$=1.0.}
	\begin{tabular}{l|c|ccc|ccc}
		\hline
		\multicolumn{1}{c|}{\multirow{3}{*}{\textbf{Method}}} & \multirow{3}{*}{\textbf{Model}} & \multicolumn{3}{c|}{\textbf{120 Clients}}                                                                                                            & \multicolumn{3}{c}{\textbf{150 Clients}}                                                                                                            \\
		\multicolumn{1}{c|}{}                                 &                                 & \multirow{2}{*}{\textbf{\begin{tabular}[c]{@{}c@{}}Maximum\\ Average UA\end{tabular}}} & \multicolumn{2}{c|}{\textbf{\begin{tabular}[c]{@{}c@{}}Comm. Overhead when\\ Reaching Average UA\end{tabular}}} & \multirow{2}{*}{\textbf{\begin{tabular}[c]{@{}c@{}}Maximum\\ Average UA\end{tabular}}} & \multicolumn{2}{c}{\textbf{\begin{tabular}[c]{@{}c@{}}Comm. Overhead when\\ Reaching Average UA\end{tabular}}} \\
		\multicolumn{1}{c|}{}                                 &                                 &                                       & \textbf{37\%}                                         & \textbf{60\%}                                        &                                       & \textbf{37\%}                                        & \textbf{60\%}                                        \\ \hline
		FedAvg                                                & \multirow{5}{*}{$A^C_6$}             & 39.06                                 & 113.24M                                               & -                                                    & 44.60                                 & 96.36M                                               & -                                                    \\
		FedAdam                                               &                                 & 27.48                                 & -                                                     & -                                                    & 39.26                                 & 356.46M                                              & -                                                    \\
		pFedMe                                                &                                 & 36.00                                 & -                                                     & -                                                    & 42.10                                 & 237.19M                                              & -                                                    \\
		MTFL                                                  &                                 & 39.20                                  & 111.21M                                               & -                                                    & 44.98                                 & 101.75M                                              & -                                                    \\ 
            \textcolor{black}{DemLearn}                                                  &                                 & 33.44                                  & -                                               & -                                                    & 31.76                                 & -                                              & -                                                    \\
  
  \hline
		FedAvg                                                & \multirow{5}{*}{$A^C_7$}             & 40.75                                 & 45.24M                                                & -                                                    & 45.06                                 & 117.99M                                              & -                                                    \\
		FedAdam                                               &                                 & 37.35                                 & 176.98M                                               & -                                                    & 39.18                                 & 444.46M                                              & -                                                    \\
		pFedMe                                                &                                 & 37.81                                 & 97.51M                                                & -                                                    & 38.58                                 & 277.98M                                              & -                                                    \\
		MTFL                                                  &                                 & 40.15                                 & 47.38M                                                & -                                                    & 45.16                                 & 110.35M                                              & -                                                    \\
            \textcolor{black}{DemLearn}                                                  &                                 & 36.02                                 & -                                                & -                                                    & 32.42                                 & -                                              & -                                                    \\
        \hline
		FedAvg                                                & \multirow{5}{*}{$A^C_8$}             & 42.80                                 & 64.45M                                                & -                                                    & 45.46                                 & 137.50M                                              & -                                                    \\
		FedAdam                                               &                                 & 40.42                                 & 249.22M                                               & -                                                    & 36.00                                 & -                                                    & -                                                    \\
		pFedMe                                                &                                 & 37.69                                 & 151.25M                                               & -                                                    & 36.39                                 & -                                                    & -                                                    \\
		MTFL                                                  &                                 & 42.52                                 & 65.74M                                                & -                                                    & 45.20                                 & 137.50M                                              & -                                                    \\ 
            \textcolor{black}{DemLearn}                                                  &                                 & 37.60                                 & 134.47M                                                & -                                                    & 36.83                                 & -                                              & -                                                    \\  \hline
		FedGKT                                                & \multirow{4}{*}{$A^C_6,A^C_7,A^C_8$}     & 61.00                                 & {0.70M}                                        & 4.97M                                                 & 64.41                                 & \textbf{0.54M}                                       & 3.72M                                                \\
		FedDKC                                                &                                 & 60.83                                 & {0.70M}                                        & 4.60M                                                & {66.89}                           & \textbf{0.54M}                                       & {\ul 2.89M}                                                \\
		\textbf{FedICT (sim)}                                 &                                 & {\ul 61.53}                   & \textbf{0.54M}                                        & {\ul 3.45M}                                       & {\ul 66.98}                        & \textbf{0.54M}                                       & \textbf{1.99M}                                          \\
		\textbf{FedICT (balance)}                             &                                 & \textbf{62.85}                           & \textbf{0.54M}                                        & \textbf{2.83M}                                       & \textbf{67.41}                                 & \textbf{0.54M}                                       & {\ul 2.89M}                                       \\ \hline
	\end{tabular}
	\label{mobile-table}
\end{table*}

\section{Ablation Study}
\subsection{Ablation Settings}
\label{ablation}
To verify that our proposed methods actually benefit from leveraging local/global data distribution information, we conduct the ablation operation $\mathcal{D}_{meta}@$ where the randomly generated data distribution vectors instead of the actual local data distribution vectors are used in FedICT.
%adopt random vectors in substitution with local data distribution vectors in FedICT, where $\mathcal{D}_{meta}$ is a base distribution for generating $\mathcal{D}_{deriv}$.
Specifically, random local data distribution vectors $d^k \sim \tau(\mathcal{D}_{meta})$, so as to simulate $d^k$ that is independent of local data distributions. According to algorithm \ref{alg2}, line 8, $d^S$ is calculated from $d^k$, so it is also set as random.
%Let $\mathcal{D}_{meta}$ denote a common continuous distribution, such as the uniform distribution $\mathcal{U}$, the normal distribution $\mathcal{N}$, the exponential distribution $\mathcal{E}$, etc, 
%\textcolor{black}{In order to let $\mathcal{D}_{deriv}$ satisfies a probability distribution}, we adopt the softmax mapping $\tau(\cdot)$ to normalize the set of samples generated from $\mathcal{D}_{meta}$, i.e. $\mathcal{D}_{meta}=\tau(\mathcal{D}_{deriv})$, so as to simulate $d^k$ that is completely independent to local data distributions.
In this paper, we try several common $\mathcal{D}_{meta}$ to generate $d^k$, which are $\mathcal{U}(0,3)$, $\mathcal{N}(0,3)$ and $\mathcal{E}(3)$. On this basis, we conduct ablation experiments with operation $\mathcal{D}_{meta}@$ on both FedICT (sim) and FedICT (balance). Specifically, both homogeneous and heterogeneous model settings are considered, with the same experimental configurations as provided in section \ref{experiments}.

\begin{table}[t]
	\caption{Average UA (\%) with different ablation operations. Results are derived on CIFAR-10 dataset, taking $\alpha$=1.0.}
	\centering
	\setlength{\tabcolsep}{10pt}
	\renewcommand\arraystretch{1.13}
\begin{tabular}{c|l|cc}
	\hline
	\multirow{5}{*}{\textbf{\begin{tabular}[c]{@{}c@{}}Model\\ Homo.\end{tabular}}}   & \multicolumn{1}{c|}{\textbf{Operation}} & \textbf{\begin{tabular}[c]{@{}c@{}}FedICT\\ (sim)\end{tabular}} & \textbf{\begin{tabular}[c]{@{}c@{}}FedICT\\ (balance)\end{tabular}} \\ \cline{2-4} 
	& $\mathcal{U}(0,3)@$                                 & 64.86                                                           & 64.63                                                               \\
	& $\mathcal{N}(0,3)@$                                 & 63.34                                                           & 64.35                                                               \\
	& $\mathcal{E}(3)@$                                 & 63.19                                                           & 63.88                                                               \\
	& \textbf{None}                           & \textbf{65.42}                                                  & \textbf{65.15}                                                      \\ \hline
	\multirow{5}{*}{\textbf{\begin{tabular}[c]{@{}c@{}}Model\\ Hetero.\end{tabular}}} & \multicolumn{1}{c|}{\textbf{Operation}} & \textbf{\begin{tabular}[c]{@{}c@{}}FedICT\\ (sim)\end{tabular}} & \textbf{\begin{tabular}[c]{@{}c@{}}FedICT\\ (balance)\end{tabular}} \\ \cline{2-4} 
	& $\mathcal{U}(0,3)@$                                 & 62.82                                                           & 62.46                                                               \\
	& $\mathcal{N}(0,3)@$                                 & 60.67                                                           & 61.75                                                               \\
	& $\mathcal{E}(3)@$                                 & 62.12                                                           & 62.47                                                               \\
	& \textbf{None}                           & \textbf{63.06}                                                  & \textbf{63.36}                                                      \\ \hline
\end{tabular}
\label{ablation-table}
\end{table}

\begin{table*}[]
\centering
	\caption{Computation complexity of existing FD methods without public datasets. Backward propagation, forward propagation, and stochastic gradient descent are denoted as BP., FP., SGD., respectively.
 }
 \renewcommand\arraystretch{1.1}
 \begin{adjustbox}{center}
\begin{tabular}{c|l|c|c|c|c}
\hline
\multirow{6}{*}{\textbf{\begin{tabular}[c]{@{}c@{}}Network\\ Termination\end{tabular}}}  & \multicolumn{1}{c|}{\textbf{Method}} & \textbf{Initialization}                & \textbf{BP./FP./SGD.}          & \textbf{Loss Computation}                                                                      & \textbf{Total}                 \\ \cline{2-6} 
                                                                                  & FedGKT                               & \multirow{3}{*}{-}            & \multirow{5}{*}{$RN^k{\rm{\cdot O}}(W^k)$} & \multirow{5}{*}{$RN^k{\rm{\cdot O}}(C)$}                                                      & \multirow{5}{*}{$RN^k{\rm{\cdot O}}(W^k)$} \\
                                                                                  & KKR-FedDKC                           &                               &                                &                                                                                   &                                \\
                                                                                  & SKR-FedDKC                           &                               &                                &                                                                                   &                                \\ \cline{2-3}
                                                                                  & \textbf{FedICT (sim)}                & \multirow{2}{*}{${\rm{O}}(N^k+C)$}   &                                &                                                                                   &                                \\
                                                                                  & \textbf{FedICT (balance)}            &                               &                                &                                                                                   &                                \\ \hline
\multirow{6}{*}{\textbf{\begin{tabular}[c]{@{}c@{}}Network\\ Edge\end{tabular}}} & \multicolumn{1}{c|}{\textbf{Method}} & \textbf{Initialization}                & \textbf{BP./FP./SGD.}          & \textbf{Loss Computation}                                                                      & \textbf{Total}                 \\ \cline{2-6} 
                                                                                  & FedGKT                               & \multirow{3}{*}{$\sum\limits_{k = 1}^K {{N^k}}{\rm{\cdot O}}(C)$}     & \multirow{5}{*}{$R\sum\limits_{k = 1}^K {{N^k}}{\rm{\cdot O}}(W^S)$}   & \multirow{2}{*}{$R\sum\limits_{k = 1}^K {{N^k}}{\rm{\cdot O}}(C)$}                                                        & \multirow{5}{*}{$R\sum\limits_{k = 1}^K {{N^k}}{\rm{\cdot O}}(W^S)$}   \\
                                                                                  & KKR-FedDKC                           &                               &                                &                                                                                   &                                \\ \cline{5-5}
                                                                                  & SKR-FedDKC                           &                               &                                & \multicolumn{1}{l|}{$R\sum\limits_{k = 1}^K {{N^k}}{\rm{\cdot O}}(C\log \frac{|\epsilon _1-\epsilon _2|}{\varepsilon})$} &                                \\ \cline{2-3} \cline{5-5}
                                                                                  & \textbf{FedICT (sim)}                & \multirow{2}{*}{$(K+\sum\limits_{k = 1}^K {{N^k}}){\rm{\cdot O}}(C)$} &                                & \multirow{2}{*}{$R\sum\limits_{k = 1}^K {{N^k}}{\rm{\cdot O}}(C)$}                                                        &                                \\
                                                                                  & \textbf{FedICT (balance)}            &                               &                                &                                                                                   &                                \\ \hline
\end{tabular}
 \end{adjustbox}
\label{computation-complex}
\end{table*}

\subsection{Results}
TABLE \ref{ablation-table} displays the experimental results with different ablation operations and model architectures. We can figure out that the average UAs of FedICTs with operation $\mathcal{D}_{meta}@$ are all degraded, regardless of adopted LKA techniques and model architecture settings.
%\textcolor{black}{This is because the randomly generated $d^k$ makes both global and local distillation ineffective, and average performances with operator $\mathcal{D}_{meta}@$ are inevitably degraded.}
%results obtained with operator $\mathcal{D}_{meta}@$ are those who replace $d^k$ with randomly generated vectors based on the given distribution. At the same time, $d^S$ is also set random since it derives from the weighting of $d^k$.
This result confirms that our methods indeed improve average user performance by transferring the knowledge of local/global data distributions. 
%\textcolor{black}{As setting $d^k$ to random make local/global data distribution vectors ineffective, the average performances with operator $\mathcal{D}_{meta}@$ are all degraded.}

%\subsection{Defects on Long-tail Overall Data}
%\section{Discussions}
\section{Analysis on Computation Cost}
We compare the computation complexity of FedICT with existing FD methods without public datasets \cite{he2020group,wu2022exploring}, as shown in TABLE \ref{computation-complex}. 
Compared with FedGKT, FedICT introduces additional computational overhead twofold: training initialization and loss computation.
At the client side, FedICT requires to compute data distribution vectors during local initialization, which introduces ${\rm{O}}(N^k+C)$ extra computation cost on client $k$ compared with previous works \cite{he2020group,wu2022exploring}. Besides, the newly introduced optimization component $J^k_{FPKD}(\cdot)$ requires additional ${RN^k\rm{\cdot O}}(C)$ computation cost.
At the server side, local data distribution vectors should be utilized to compute the global data distribution vector during global initialization, where additional $K{\rm{\cdot O}}(C)$ computational cost is required. Likewise, $J^k_{LKA}(\cdot)$ introduced by LKA needs extra ${R\sum\limits_{k = 1}^K {{N^k}}\rm{\cdot O}}(C)$ computation in the server, regardless of similarity-based or class-balanced technique is adopted.

Although extra computation cost is introduced during initialization and each training round, we still suggest that FedICT is a computation-efficient FD paradigm compared with prior works \cite{he2020group,wu2022exploring}. On the one hand, the additional computation cost introduced during initialization and loss computation is orders of magnitude less than forward/backward propagation or gradient descent, i.e. ${\rm{O}}(N^k+C) \ll {N^k\rm{\cdot O}}(W^k)$, $K{\rm{\cdot O}}(C) \ll {\sum\limits_{k = 1}^K {{N^k}}\rm{\cdot O}}(W^S)$ during initialization and ${RN^k\rm{\cdot O}}(C) \ll {RN^k\rm{\cdot O}}(W^k)$, $RK{\rm{\cdot O}}(C) \ll {R\sum\limits_{k = 1}^K {{N^k}}\rm{\cdot O}}(W^S)$ during model training.
%$\frac{{{\rm{O}}({N^k} + C)}}{R} + {N^k}{\rm{O}}(C)={N^k} \cdot {\rm{O}}(\frac{1}{R} + \frac{C}{{R{N^k}}} + C) \ll {N^k\rm{\cdot O}}(W^k)$
%at client $k$, and $(\frac{K}{R} + \sum\limits_{k = 1}^K {{N^k}} ) \cdot {\rm{O}}(C) \ll {\sum\limits_{k = 1}^K {{N^k}}\rm{\cdot O}}(W^S)$ on the server. 
On the other hand, the overall computational overhead is proportional to the number of training rounds, and FedICT can effectively accelerate model convergence with at least 25\% and 14\% fewer training rounds to achieve the same average UA compared with FedGKT and FedDKC, respectively, as discussed in section \ref{convergence-analysis}. 
Therefore, we can conclude that FedICT generally requires less computation cost than state-of-the-art methods.
%Thus, we can conclude that FedICT requires less computational cost than state-of-the-art.

%\subsection{Extensions of FedICT}
%The main framework (FedICT) and feasible techniques (FPKD, LKA) are provided to tackle communication, model heterogeneity and multi-task issues for MEC. Nevertheless, federated multi-task distillation is a multi-dimensional problem that requires multifaceted technical and performance trade-offs; system extensions should incorporate common techniques in federated edge learning, such as device selection, resource allocation, asynchronous optimization, and privacy protection \cite{tak2020federated,lim2020federated,yu2021toward}. In this way can make more practical optimization for federated multi-task distillation in MEC, and provide fundamental technical support for its wide application in real scenes.

\section{Conclusion}
\label{conclusion}
This paper proposes a federated multi-task distillation framework for multi-access edge computing (FedICT). In our framework, local and global knowledge is disaffected to achieve client-side adaptation to multiple tasks while alleviating client drift derived from divergent client-side optimization directions. Specifically, we propose FPKD and LKA techniques to reinforce the clients' fitting to local data or to match the transferred local knowledge to better suit generalized representation. To our best knowledge, this paper is the first work that enables federated multi-task learning to be deployed practically in multi-access edge computing. Extensive experiments on both image classification and transportation mode detection demonstrate that our proposed methods achieve superior performance than the state-of-the-art while improving communication efficiency and convergence speed by a large margin without requiring additional public datasets.

\section*{Acknowledgments}
We thank Hui Jiang, Qingxiang Liu and Xujing Li from Institute of Computing Technology, Chinese Academy of Sciences, Jinda Lu from University of Science and Technology of China, Zhiqi Ge from Zhejiang University, Zixuan Li from Sun Yat-sen University and Yiming Cheng from University of the Arts London for inspiring suggestions.
\section*{Acknowledgment}
\ifCLASSOPTIONcaptionsoff
  \newpage
\fi

\bibliographystyle{IEEEtran}
% \bibliography{ref.bib}

\vspace{-33pt}
% \begin{IEEEbiography}
% 	%图片
% \end{IEEEbiography}
\begin{IEEEbiography}[{\includegraphics[width=1in,height=1.25in,clip,keepaspectratio]{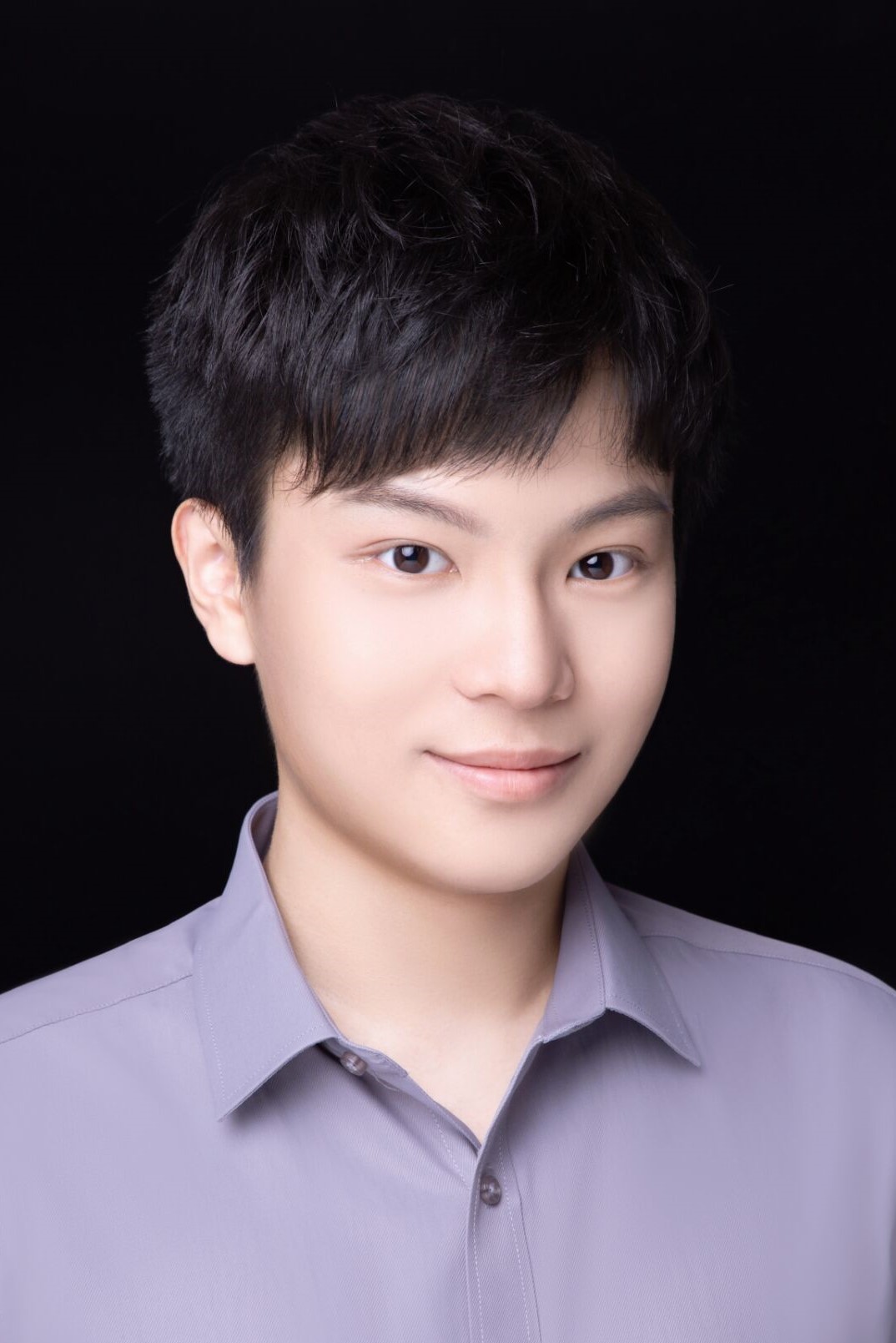}}]{Zhiyuan Wu}
	(Member, IEEE) is currently a research assistant with the Institute of Computing Technology, Chinese Academy of Sciences. He is also a member of Distributed Computing and Systems Committee as well as the Artificial Intelligence and Pattern Recognition Committee in China Computer Federation (CCF). His research interests include mobile edge computing, federated learning, and distributed systems.
\end{IEEEbiography}

\begin{IEEEbiography}[{\includegraphics[width=1in,height=1.25in,clip,keepaspectratio]{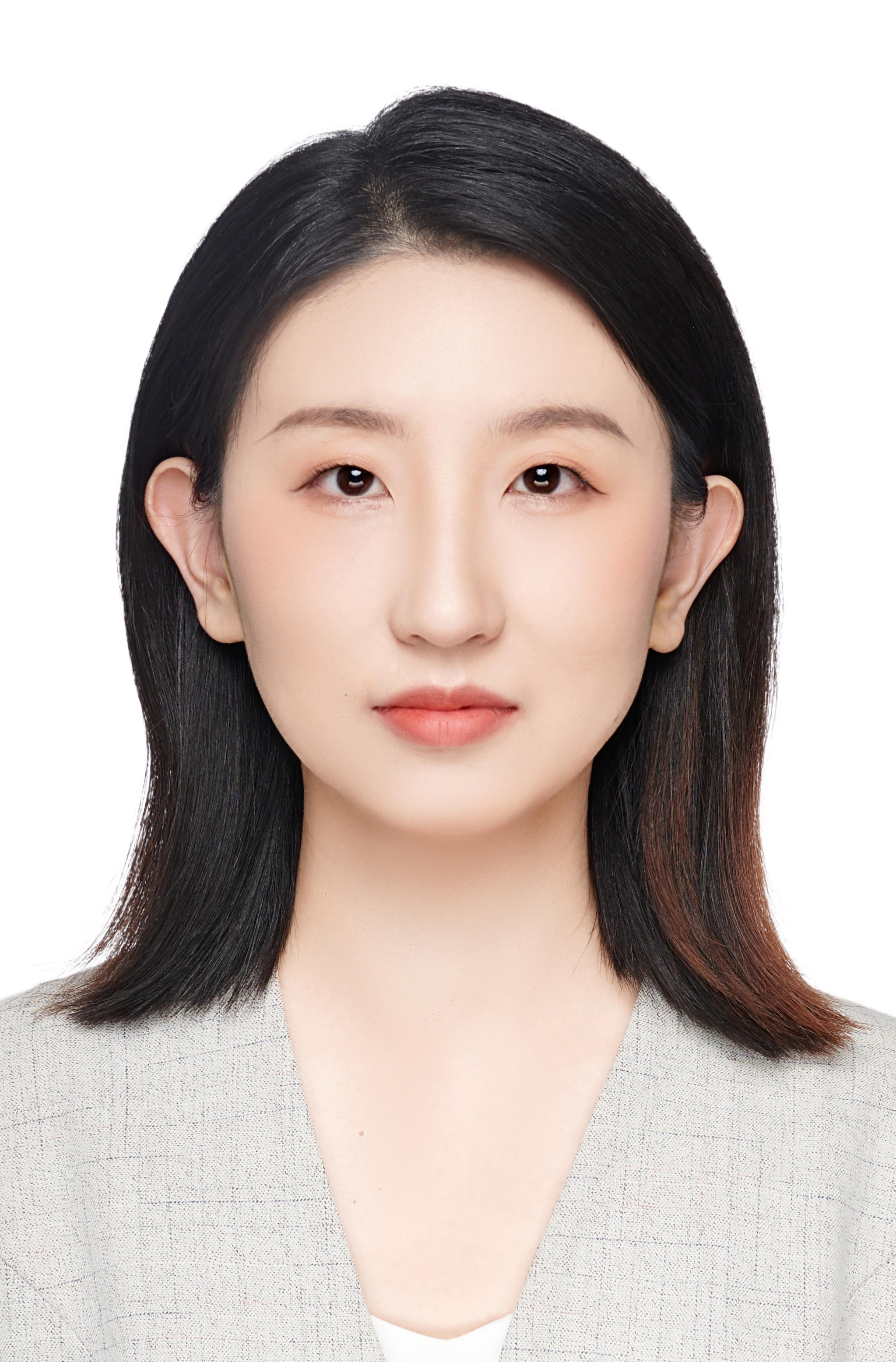}}]{Sheng Sun}
	received her B.S. and Ph.D degrees in computer science from Beihang University, China, and the University of Chinese Academy of Sciences, China, respectively. She is currently an assistant professor at the Institute of Computing Technology, Chinese Academy of Sciences, Beijing, China. Her current research interests include federated learning, mobile computing and edge intelligence. 
\end{IEEEbiography}

\begin{IEEEbiography}[{\includegraphics[width=1in,height=1.25in,clip,keepaspectratio]{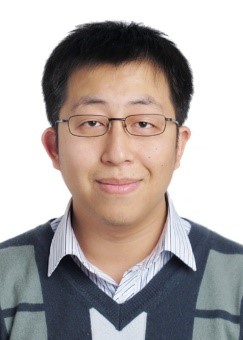}}]{Yuwei Wang}
	(Member, IEEE) received his Ph.D. degree in computer science from the University of Chinese Academy of Sciences, Beijing, China. He is currently an associate professor at the Institute of Computing Technology, Chinese Academy of Sciences. He has been responsible for setting over 30 international and national standards, and also holds various positions in both international and national industrial standards development organizations (SDOs) as well as local research institutions, including the associate rapporteur at the ITU-T SG16 Q5, and the deputy director of China Communications Standards Association (CCSA) TC1 WG1. His current research interests include federated learning, mobile edge computing, and next-generation network architecture.
\end{IEEEbiography}

\begin{IEEEbiography}[{\includegraphics[width=1in,height=1.25in,clip,keepaspectratio]{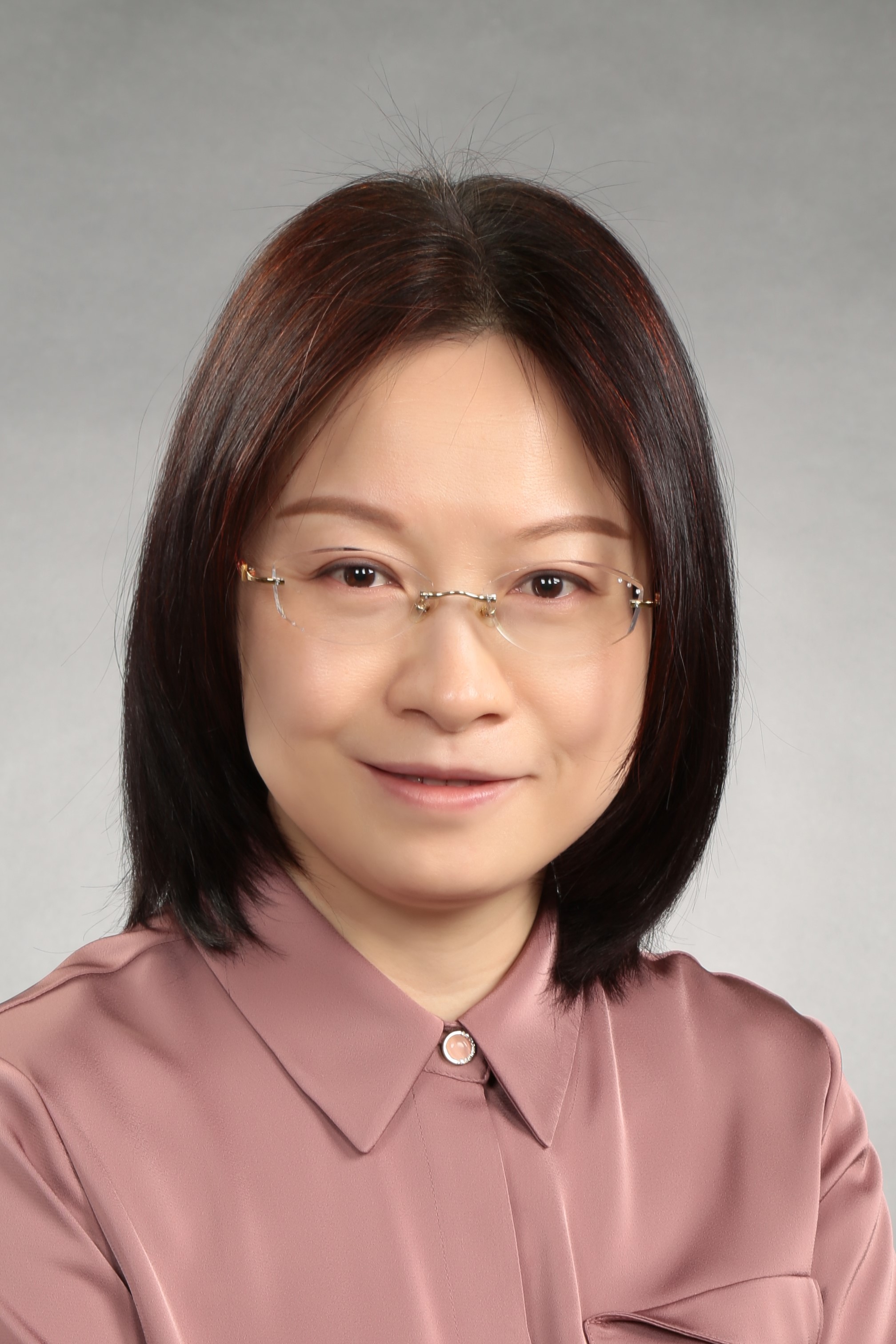}}]{Min Liu}
	(Senior Member, IEEE) received her Ph.D degree in computer science from the Graduate University of the Chinese Academy of Sciences, China. Before that, she received her B.S. and M.S. degrees in computer science from Xi’an Jiaotong University, China. She is currently a professor at the Institute of Computing Technology, Chinese Academy of Sciences, and also holds a position at the Zhongguancun Laboratory. Her current research interests include mobile computing and edge intelligence.
\end{IEEEbiography}

\begin{IEEEbiography}[{\includegraphics[width=1in,height=1.25in,clip,keepaspectratio]{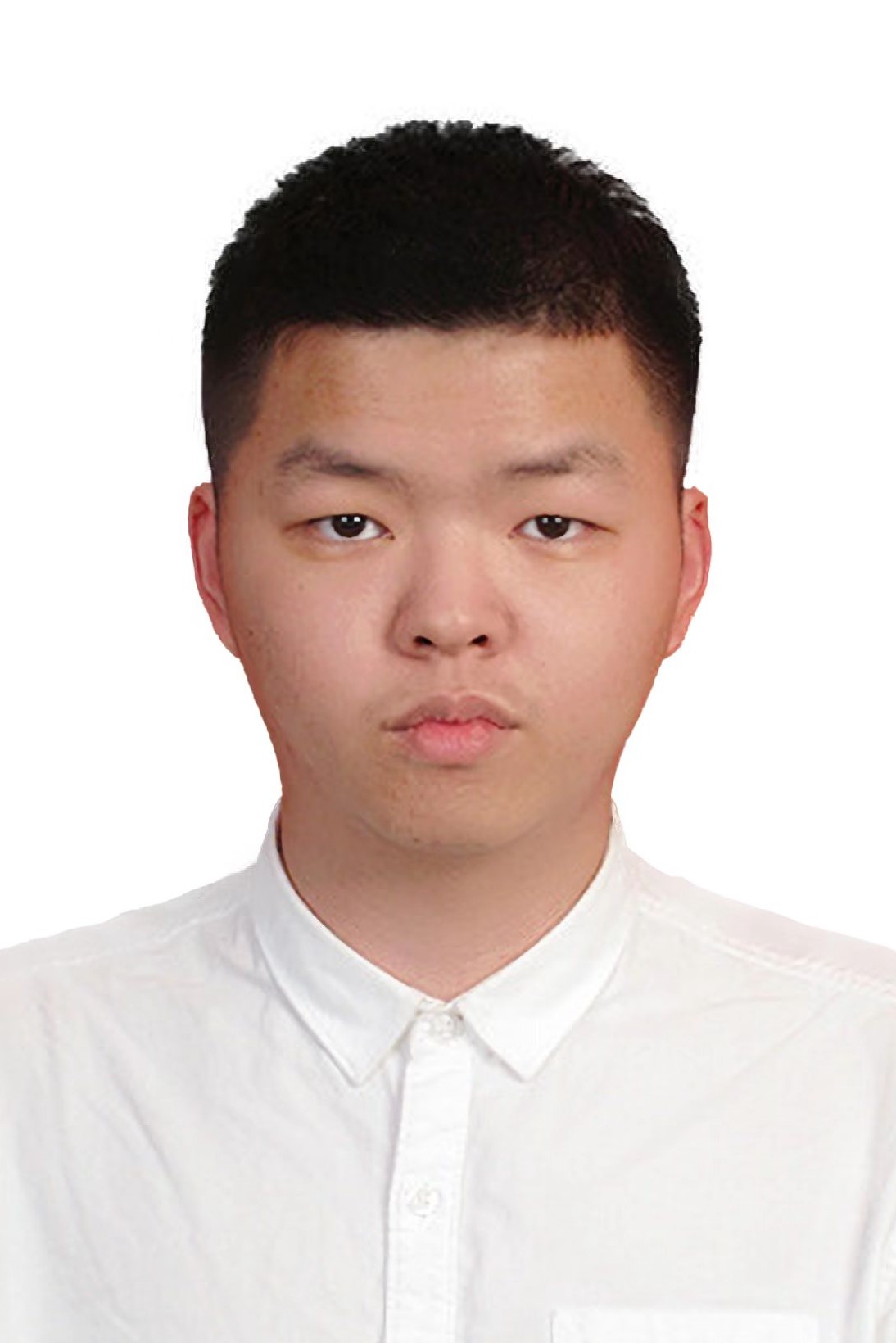}}]{Quyang Pan}
	is currently a master’s candidate with the Institute of Computing Technology, Chinese Academy of Sciences. He is an outstanding competitive programmer who has won several gold medals in international and national contests such as ACM-ICPC, CCF-CCSP, etc. His research interests include federated learning and reinforcement learning.
\end{IEEEbiography}

\begin{IEEEbiography}[{\includegraphics[width=1in,height=1.25in,clip,keepaspectratio]{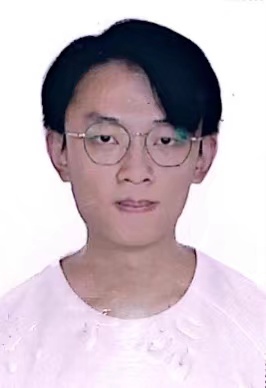}}]{Xuefeng Jiang}
	is currently a Ph.D candidate with the Institute of Computing Technology, Chinese Academy of Sciences. Before that, he received his bachelor degree with honor at Beijing University of Posts and Telecommunications. His research interests include distributed optimization and machine learning.
\end{IEEEbiography}

\begin{IEEEbiography}[{\includegraphics[width=1in,height=1.25in,clip,keepaspectratio]{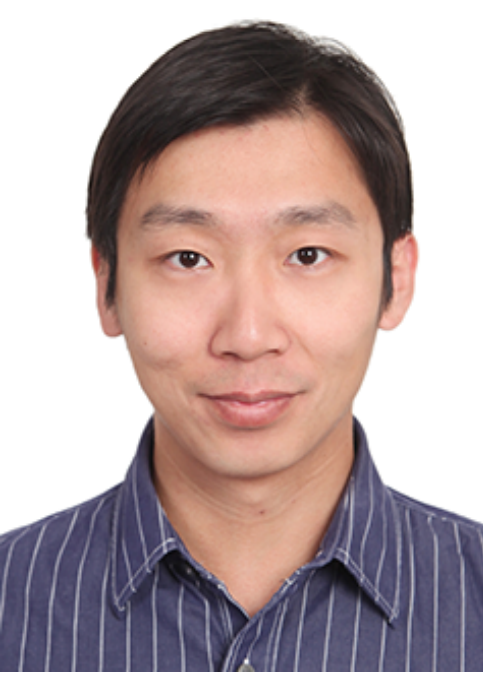}}]{Bo Gao} (Member, IEEE)
	received his M.S. degree in electrical engineering from the School of Electronic Information and Electrical Engineering at Shanghai Jiaotong University, Shanghai, China in 2009, and his Ph.D. degree in computer engineering from the Bradley Department of Electrical and Computer Engineering at Virginia Tech, Blacksburg, USA in 2014. He was an Assistant Professor with the Institute of Computing Technology at Chinese Academy of Sciences, Beijing, China from 2014 to 2017. He was a Visiting Researcher with the School of Computing and Communications at Lancaster University, Lancaster, UK from 2018 to 2019. He is currently an Associate Professor with the School of Computer and Information Technology at Beijing Jiaotong University, Beijing, China. He has directed a number of research projects sponsored by the National Natural Science Foundation of China (NSFC) or other funding agencies. He is a member of IEEE, ACM, and China Computer Federation (CCF). His research interests include wireless networking, mobile/edge computing, multiagent systems, and machine learning.
\end{IEEEbiography}
\vfill
\newpage
\appendices
\onecolumn
\end{document}